\documentclass{article}

\usepackage{microtype}
\usepackage{graphicx}
\usepackage{booktabs} 

\usepackage{hyperref}

\usepackage[accepted]{icml2020}

\usepackage{diagbox}
\usepackage{adjustbox}

\usepackage{subcaption}

\usepackage[font=footnotesize, labelfont=bf]{caption}

\usepackage{algorithm, algorithmic}

\usepackage{amsmath,amsfonts,bm}

\def\eqref#1{equation~\ref{#1}}

\def\1{\bm{1}}

\DeclareMathAlphabet{\mathsfit}{\encodingdefault}{\sfdefault}{m}{sl}
\SetMathAlphabet{\mathsfit}{bold}{\encodingdefault}{\sfdefault}{bx}{n}

\newcommand{\R}{\mathbb{R}}

\usepackage{bbm}

\newcommand{\ie}{\textit{i.e., }}
\newcommand{\eg}{\textit{e.g., }}

\usepackage{color}

\newcommand{\scomment}[1]{}

\usepackage{xspace}
\newcommand{\name}{TGT\xspace}
\newcommand{\abr}{GCE\xspace}
\newcommand{\abrs}{GCEs\xspace}
\newcommand{\base}{DBM\xspace}

\icmltitlerunning{Explaining Low-Dimensional Representations}

\begin{document}

\twocolumn[
\icmltitle{Explaining Groups of Points in Low-Dimensional Representations}



\icmlsetsymbol{equal}{*}

\begin{icmlauthorlist}
	\icmlauthor{Gregory Plumb}{cmu}
	\icmlauthor{Jonathan Terhorst}{um}
	\icmlauthor{Sriram Sankararaman}{ucla}
	\icmlauthor{Ameet Talwalkar}{cmu,dai}
\end{icmlauthorlist}

\icmlaffiliation{cmu}{Carnegie Mellon University}
\icmlaffiliation{um}{University of Michigan}
\icmlaffiliation{ucla}{University of California, Los Angeles}
\icmlaffiliation{dai}{Determined AI}

\icmlcorrespondingauthor{Gregory Plumb}{gdplumb@andrew.cmu.edu}

\icmlkeywords{Machine Learning, ICML, Interpretable ML, Counterfactual, Global Explanation}

\vskip 0.3in
]



\printAffiliationsAndNotice{}  

\begin{abstract}
	A common workflow in data exploration is to learn a low-dimensional representation of the data, identify groups of points in that representation, and examine the differences between the groups to determine what they represent.  
	We treat this workflow as an interpretable machine learning problem by leveraging the model that learned the low-dimensional representation to help identify the key differences between the groups.  
	To solve this problem, we introduce a new type of explanation, a Global Counterfactual Explanation (\abr), and our algorithm, Transitive Global Translations (\name), for computing \abrs.   
	\name identifies the differences between each pair of groups using compressed sensing but constrains those pairwise differences to be consistent among all of the groups.  
	Empirically, we demonstrate that \name is able to identify explanations that accurately explain the model while being relatively sparse, and that these explanations match real patterns in the data.  
\end{abstract}

\section{Introduction}

A common workflow in data exploration is to: 1) learn a low-dimensional representation of the data, 2) identify groups of points (\ie clusters) that are similar to each other in that representation, and 3) examine the differences between the groups to determine what they represent.  
We focus on the third step of this process: answering the question \textit{``What are the key differences between the groups?''} 

\begin{figure}[t]
	\centering
	\includegraphics[width =  \linewidth]{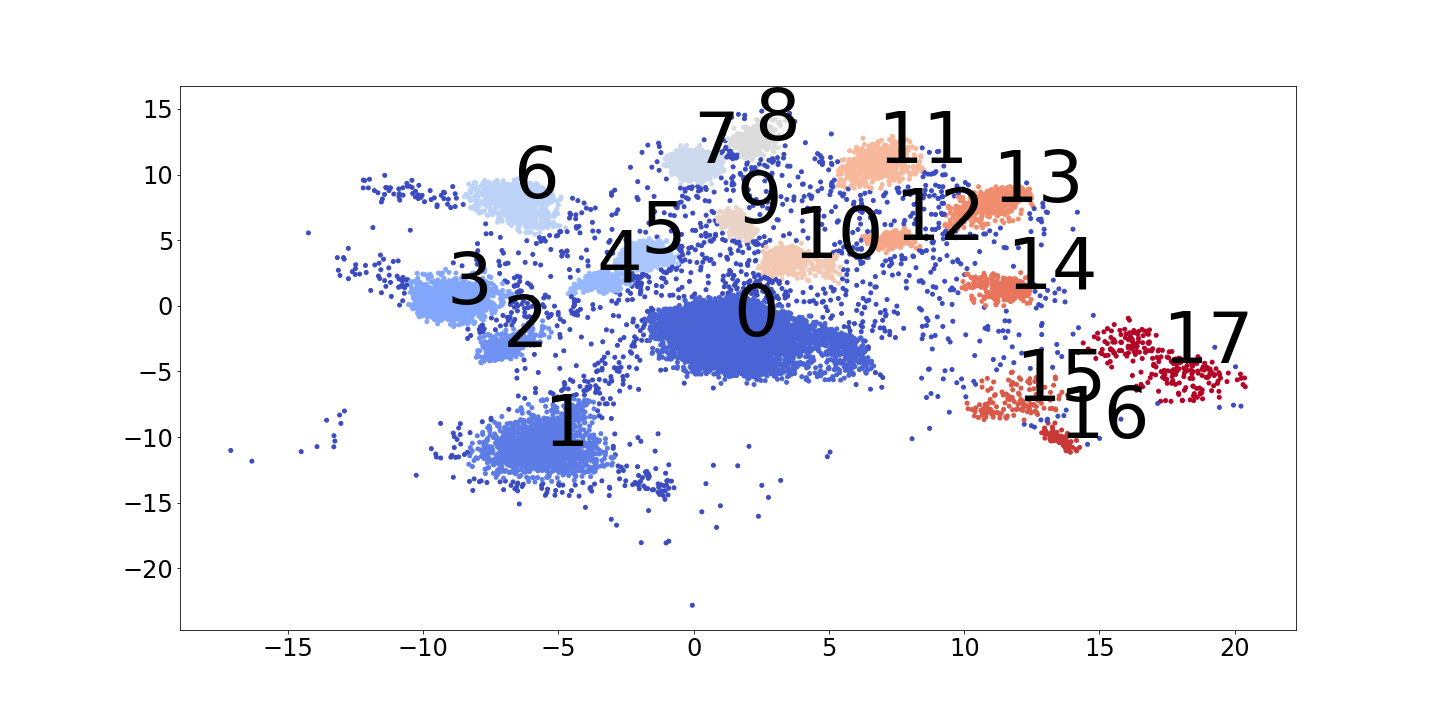}
	\caption{\label{fig:bipolar-rep}  A representation learned for a single-cell RNA sequence dataset using the model from \citep{ding2018interpretable}.  
		Previous work on this dataset showed that these groups of cells correspond to different cell-types \citep{shekhar2016comprehensive}.  
		The goal of a \abr is to use this representation  to identify the changes in gene expression that are associated with a change of cell type.}
\end{figure}

For data exploration, this is an interesting question because the groups often correspond to an unobserved concept of interest and, by identifying which features differentiate the groups, we can learn something about that concept of interest.  
For example, consider single-cell RNA analysis.
These datasets measure the expression levels of many genes for sampled cells.
Usually the cell-type of each of those cells is unknown. 
Because gene expression and cell-type are closely related, the groups of points that can be seen in a low-dimensional representation of the dataset often correspond to different cell-types (Figure \ref{fig:bipolar-rep}).  
By determining which gene expressions differentiate the groups, we can learn something about the connection between gene expression and cell-type.  

\begin{figure}[t]
	\centering
	\includegraphics[width =  \linewidth]{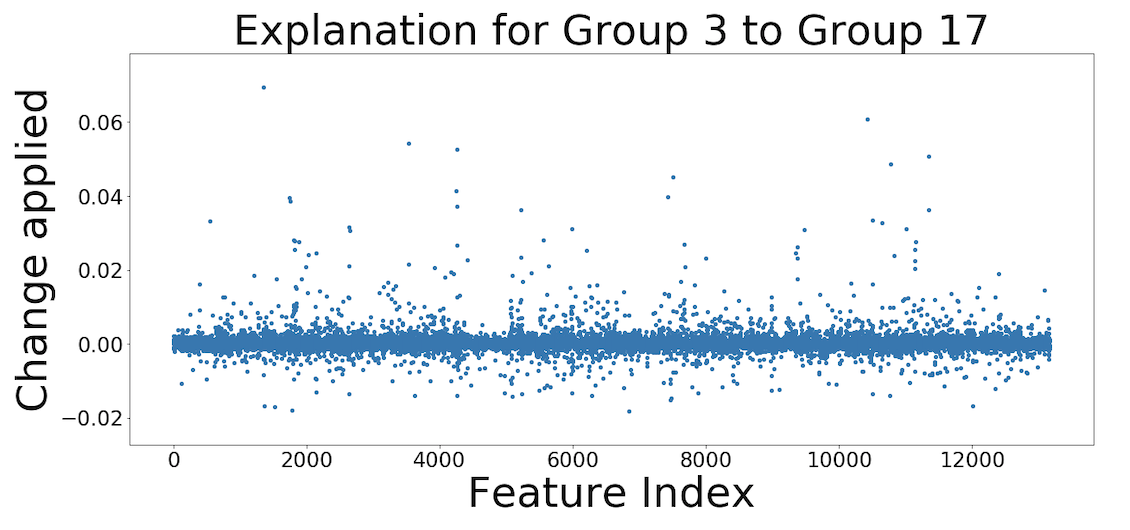} 
	\caption{\base's explanation for the difference in gene expression between the cells in Group 3 and Group 17.  
		The x-axis shows which feature index (gene expression) is being changed and the y-axis shows by how much.  
		Because it is very high dimensional and not sparse, it is difficult to use \base to determine what the key differences actually are between this pair of groups (\ie which genes differentiate these cell-types). }
	\label{fig:bipolar-dbm-example}  
	\includegraphics[width =  \linewidth]{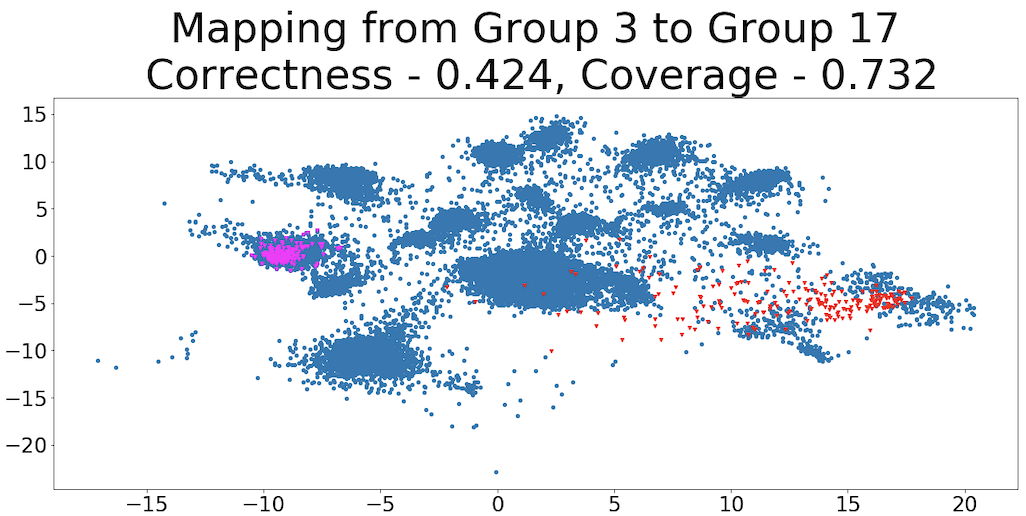} 
	\caption{By thresholding \base's explanation for the differences  between Group 3 and Group 17 to include only the $k$ largest changes (250 in this case), we can make it sparse enough to be human interpretable.  
		However, this simplified explanation is no longer an effective explanation.  
		We show this visually: the magenta points are the representations of points sampled randomly from Group 3 and the red points are the representations of those points after the explanation was applied to them.  
		We can see that the red points are usually not in Group 17 (poor correctness) and that they do not cover much of Group 17 (poor coverage).
		These metrics will be defined in Section \ref{sec:method}.}
	\label{fig:bipolar-dbm-example-applied}  
\end{figure}

One common approach for answering this question is manual interpretation. 
One simple way to do this, that we will use as a naive baseline, is to calculate the Difference Between the Mean (\base) value of each feature in the original input space between a pair of groups. 
For example, consider using \base to explain the differences between the cells in Group 3 and Group 17 from Figure \ref{fig:bipolar-rep}.  
In this case,  \base's explanation contains many hundreds of non-zero elements, which is far too many to be understood by a person (Figure \ref{fig:bipolar-dbm-example}). 
 If we make \base's explanation sparse by thresholding it to include only the $k$ largest changes, it is no longer an effective explanation because it no longer reliably maps points from Group 3 to Group 17 (Figure \ref{fig:bipolar-dbm-example-applied}).  
More generally, manual interpretation can be time-consuming and typically ad-hoc, requiring the analyst to make arbitrary decisions that may not be supported by the data. 

Another, more principled, method is statistical hypothesis testing for the differences between features across groups \citep{shaffer1995multiple}.
However, the trade-off between the power of these tests and their false positive rate becomes problematic in high-dimensional settings.  

Both manual interpretation and statistical testing have an additional key shortcoming: they do not make use of the model that learned the low dimensional representation that was used to define the groups in the first place. 
Intuitively, we would expect that, by inspecting this model directly, we should be able to gain additional insight into the patterns that define the groups. 
With this perspective, answering our question of interest becomes an interpretable machine learning problem.  
Although there are a wide variety of methods developed in this area \cite{ribeiro2016should, lundberg2017unified, wang2015falling, caruana2015intelligible, ribeiro2018anchors, zhang2018interpreting}, none of them are designed to answer our question of interest.
See Section \ref{sec:RelatedWork} for further discussion. 

To answer our question of interest, we want a  \textit{counterfactual} explanation because our goal is to identify the key differences between Group A and Group B using the low-dimensional representation and the most natural way to do this is to find a transformation that causes the model to assign transformed points from Group A to Group B.  
Additionally, we want a \textit{global}  explanation because we want to find a explanation that works for all of the points in Group A and because we want the complete set of explanations to be consistent (\ie symmetrical and transitive) among all the groups.
See Section \ref{sec:alg} for further discussion of our definition of consistency in this context.  
Hence, our goal is to find a Global Counterfactual Explanation (\abr).  
Although the space of possible transformations is very large,  we consider translations in this work because their interpretability can be easily measured using sparsity.  

\textbf{Contributions:}
To the best of our knowledge, this is the first work that explores \abrs.  
Motivated by the desire to generate a simple (\ie sparse) explanation between each pair of groups, we derive an algorithm to find these explanations that is motivated by compressed sensing  \cite{tsaig2006extensions, candes2006compressive}. 
However, the solutions from compressed sensing are only able to explain the differences between one pair of groups.  
As a result, we generalize the compressed sensing solution to find a set of consistent explanations among all groups simultaneously.  
We call this algorithm Transitive Global Translations (\name).  

We demonstrate the usefulness of \name with a series of experiments on synthetic, UCI, and single-cell RNA datasets.  
In our experiments, we measure the effectiveness of explanations using correctness and coverage, with sparsity as a proxy metric for interpretability, and we compare the patterns the explanations find to those we expect to be in the data. 
We find the \name clearly outperforms \base at producing sparse explanations of the model and that its explanations match domain knowledge.  
\footnote{Code for all algorithms and experiments is available at \href{https://github.com/GDPlumb/ELDR}{https://github.com/GDPlumb/ELDR}}

\section{Related Work}\label{sec:RelatedWork}

Most of the literature on cluster analysis focuses on defining the clusters; the interpretation methods discussed in that literature are primarily manual inspection/visualization or statistical testing \cite{jiang2004cluster}.  
Consequently, the focus of our related work will be on interpretable machine learning.  
Although interpretability is often loosely defined and context specific \cite{lipton2016mythos},  we categorize existing methods along two axes in order to demonstrate how a \abr differs from them.  
Those axes are the explanation's \textit{level} and its \textit{form}. 

The first axis used to categorize explanations is their level:  local or global.  
A \textit{local} explanation explains a single prediction made by the model \cite{ribeiro2016should, lundberg2017unified, plumb2018model}.
\citet{kauffmann2019clustering} studies a problem closely related to ours of explaining why a point was assigned to its cluster/group.  
A \textit{global} explanation will explain multiple predictions or the entire model at once \cite{wang2015falling, caruana2015intelligible, ribeiro2018anchors}.  

The second axis we use to categorize explanations is their form:  feature attribution, approximation, or counterfactual.  
A \textit{feature attribution} explanation assigns a value measuring how each feature contributed to the model's prediction(s) \cite{lundberg2017unified, sundararajan2017axiomatic}.  
Importantly, it is necessary to define a baseline value for the features in order to compute these explanations. 
An \textit{approximation} explanation approximates the model being explained using a function that is simple enough to be considered directly interpretable (\eg a sparse linear model or a small decision tree) across some neighborhood, which could be centered around a point or could be the entire input space \cite{ribeiro2016should, plumb2018model, wang2015falling, caruana2015intelligible, ribeiro2018anchors}.  
A \textit{counterfactual} explanation finds a transformation of the input(s) such that the transformed version of the input is treated in a specific way by the model \cite{zhang2018interpreting, dhurandhar2018explanations, goyal2019counterfactual, dhur2019model}. 

For various reasons, it would be challenging to use other types of explanations to construct a \abr.  
Local explanations would have to be aggregated in order to produce an explanation that applies to a group of points and it would be nontrivial to ensure that the resulting ``aggregated group explanations'' are consistent (\ie symmetrical and transitive).  
For feature attribution and local approximation explanations, it is difficult to guarantee that the baseline value or neighborhood they consider is defined broadly enough to find the transformation we want.  
For global approximation explanations, we might not be able to approximate a complex model well enough to find the transformation we want because of the accuracy-interpretability trade-off that stems from the complexity constraint on the explanation model \cite{lipton2016mythos}.  
For a concrete example of these difficulties, see the Appendix \ref{appendix:rw}.
This example uses Integrated Gradients \cite{sundararajan2017axiomatic} which is a local feature attribution method that produces symmetrical and transitive explanations with respect to a single class.


\section{Global Counterfactual Explanations}\label{sec:method}

We will start by introducing our notation, more formally stating the goal of a \abr, and defining the metrics that we will use to measure the quality of \abrs.
We do this under the simplifying assumption that we have only two groups of points that we are interested in.  
Then, in Section \ref{sec:cs}, we will demonstrate the connection between finding a \abr and compressed sensing.
We use that connection to derive a loss function we can minimize to find a \abr between a single pair of groups of points. 
Finally, in Section \ref{sec:alg}, we will remove our simplifying assumption and introduce our algorithm, \name, for finding a set of consistent \abrs among multiple groups of points.  

\textbf{Notation:}
Let $r: \R^d \rightarrow \R^m$ denote the function that maps the points in the feature space into a lower-dimensional representation space.
The only restriction that we place on $r$ is that it is differentiable (see the Appendix \ref{appendix:function} for more discussion on $r$).   
Suppose that we have two regions of interest in this representation:  $R_{initial}, R_{target} \subset \R^m$.
Let $X_{initial}$ and $X_{target}$ denote their pre-images.   
Then, our goal is to find the key differences between $X_{initial}$ and $X_{target}$ in $\R^d$ and, unlike manual interpretation or statistical testing, we will treat this as an interpretable machine learning problem by using $r$ to help find those key differences. 

\textbf{Defining the Goal of GCEs:}
At a high level, the goal of a \abr is to find a transformation that takes the points in $X_{initial}$ and transforms them so that they are mapped to $R_{target}$ by $r$; in other words, $r$ treats the transformed points from $X_{initial}$ as if they were points from $X_{target}$.
Formally, the goal is to find a transformation function $t:\R^d \rightarrow \R^d$ such that:
\begin{equation} \label{eq:goal}
    r(t(x)) \in R_{target} \ \forall x \in X_{initial} 
\end{equation}
Because we are using $t$ as an explanation, it should be as simple as possible.
Since they are very simple and their complexity can be readily measured by their sparsity, we limit $t$ to a translation:  
\begin{equation} \label{eq:t}
    t(x) = x + \delta
\end{equation}

\textbf{Measuring the Quality of \abrs:}
To measure the quality of a \abr we use two metrics: \textit{correctness} and \textit{coverage}.  
Correctness measures the fraction of points mapped from $X_{initial}$ into $R_{target}$.
Coverage measures the fraction of points in $R_{target}$ that transformed points from $X_{initial}$ are similar to.
Mathematically, we define correctness as:
\begin{equation}\label{eq:cor}
\resizebox{0.9 \linewidth}{!} 
{
    $cr(t) = \frac{1}{|X_{initial}|} \sum\limits_{x \in X_{initial}} \mathbbm{1}[\exists x' \in X_{target} \ | \ ||r(t(x)) - r(x')||_2^2 \le \epsilon]$
}   
\end{equation}
And coverage as:
\begin{equation} \label{eq:cov}
\resizebox{0.9 \linewidth}{!} 
{
    $cv(t) = \frac{1}{|X_{target}|} \sum\limits_{x \in X_{target}} \mathbbm{1}[\exists x' \in X_{initial} \ | \ ||r(x) - r(t(x'))||_2^2 \le \epsilon]$
}
\end{equation}

Clearly, correctness is a necessary property because an explanation with poor correctness has failed to map points from $X_{initial}$ into $R_{target}$ (Equation \ref{eq:goal}).  
However, coverage is also a desirable property because, intuitively, an explanation with good coverage has captured all of the differences between the groups.  

Defining these metrics requires that we pick a value of $\epsilon$.\footnote{%
We use an indicator based on $l_2$ distance and the points themselves for two reasons.  
First, it is necessary for the definition of coverage.  
Second, it makes it easier to define  $R_{target}$ for representations that are more than two-dimensional.}
Observe that, if $X_{initial} = X_{target}$ and $t(x) = x$, then $cr(t) = cv(t)$ and we have a measure of how similar a group of points is to itself.  
After $r$ has been learned, we increase $\epsilon$ until this self-similarity metric for each group of points in the learned representation is between 0.95 and 1.   

\begin{figure}[h]
	\centering 
	\includegraphics[width =0.5 \linewidth]{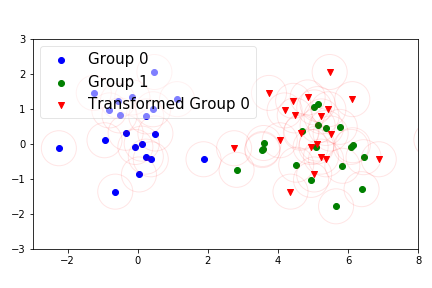}
	\caption{Two groups of points and the transformed version of Group 0.  
		The red circles indicate the balls of radius $\epsilon$ used to calculated the metrics.
		Observe that the translation has both good correctness and coverage.}
	\label{fig:3-good}
	\includegraphics[width = 0.5\linewidth]{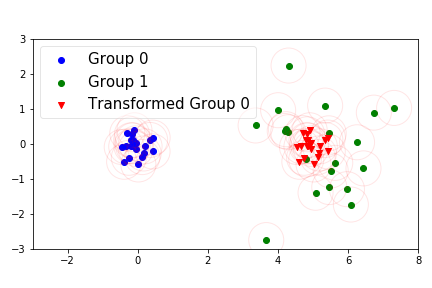}
	\caption{The same idea as Figure \ref{fig:3-good}, but now Group 0 has a smaller variance than Group 1.
		Observe that the translation has good correctness but poor coverage.}
	\label{fig:3-bad}
	\includegraphics[width = 0.5\linewidth]{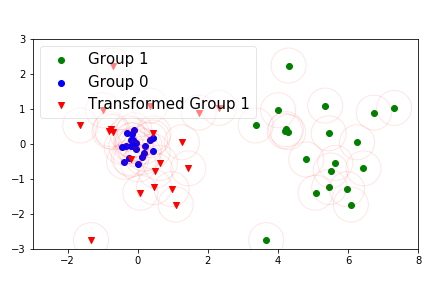}
	\caption{The same setup as in Figure \ref{fig:3-bad}, but showing what happens when the negative of the translation is applied to Group 1.
		Observe that it has good coverage but poor correctness.}
	\label{fig:3-reverse}
\end{figure}

\textbf{A Simple Illustration:}
We will now conclude our introduction to \abrs with a simple example to visualize the transformation function and the metrics.  
In  Figures \ref{fig:3-good}, \ref{fig:3-bad}, and \ref{fig:3-reverse}, the data is generated from two Gaussian distributions with different means and $r(x) = x$.
We use \base between Group 1 and Group 0 to define the translation/explanation.  
In Figure \ref{fig:3-good}, the two distributions have an equal variance and, as a result, the translation is an effective explanation with good correctness and coverage.  
In Figures \ref{fig:3-bad} and \ref{fig:3-reverse}, Group 0 has a smaller variance than Group 1.
Because a simple translation cannot capture that information\footnote{%
This is true regardless of how the translation is found (\eg \name or \base).
So this example also motivates the usefulness of more complex transformation functions.}%
, the translation has poor coverage from Group 0 to Group 1 while its negative has poor correctness from Group 1 to Group 0.  
This illustrates the connection between correctness and coverage that we will discuss more in Section \ref{sec:alg}.  

\subsection{Relating \abrs and Compressed Sensing} \label{sec:cs}
We will now demonstrate how the problem of finding a \abr between a pair of groups is connected to compressed sensing.
We start with an ``ideal'' loss function that is too difficult to optimize and then make several relaxations to it.  

In principle, Equations \ref{eq:goal}, \ref{eq:cor}, or \ref{eq:cov} define objective functions that we could optimize, but they are discontinuous and hence difficult to optimize.
To progress towards a tractable objective, first consider a continuous approximation of correctness (Equation \ref{eq:cor}):  
\begin{equation}
\resizebox{0.9 \linewidth}{!} 
{
    $loss_{min}(t) = \frac{1}{|X_{initial}|} \sum\limits_{x \in X_{initial}} \min\limits_{x' \in X_{target}} ||r(t(x)) - r(x')||_2^2$
}
\end{equation}
This loss could be optimized by gradient descent using auto-differentiation software, although doing so might be difficult because of the $min$ operation.  
Consequently, we consider a simplified version of it:
\begin{equation}
\resizebox{0.9 \linewidth}{!} 
{
	$loss_{mean}(t) = \frac{1}{|X_{initial}|} \sum\limits_{x \in X_{initial}} ||r(t(x)) - \bar r_{target}||_2^2$
}
\end{equation}
where $\bar r_{i} = \frac{1}{|X_{i}|} \sum\limits_{x \in X_{i}} r(x)$.  \footnote{%
Note that $loss_{min}$ clearly optimizes for correctness and it does not directly penalize coverage.  
However, because $loss_{mean}$  encourages $r(t(x))$ to be close to $\bar r_{target}$, it is optimizing for correctness at the expense of coverage.  
In Section \ref{sec:alg}, we will discuss why this is not as harmful as it seems in our setting where $t$ is a symmetrical translation.} 

Next, we are going to make use of our restriction of $t$ to a translation, $\delta$, and assume, for now, that $r$ is a linear mapping: $r(x) = Ax$ for $A \in \R^{m \times d}$.  
Although this assumption about $r$ will not be true in general, it allows us to connect a \abr to the classical compressed sensing formulation.  
Under these constraints, we have that:
\begin{equation}
\resizebox{0.9\linewidth}{!}{$loss_{mean+linear}(\delta) =  \frac{1}{|X_{initial}|} \sum\limits_{x \in X_{initial}} ||A(x + \delta) - \bar r_{target}||_2^2 $} \label{eq:restrict}
\end{equation}
By setting its derivative to zero,  we find that solving $A\delta = \bar r_{target} - \bar r_{initial}$ yields an optimal solution to Equation \ref{eq:restrict}.  
Observe that this is an undetermined linear system, because $m < d$, and so we can always find such a $\delta$.  

Recall that, in general, we want to find a sparse $\delta$.  
This is partially because we need the explanation to be simple enough to be understood by a person, but also because the explanation is hopefully modeling real world phenomena and these phenomena often have nice structure (\eg sparsity).  
Then, because we are finding a \abr by solving $A\delta = \bar r_{target} - \bar r_{initial}$\footnote{%
Note that \base solves this but does not consider sparsity.}
, we can see that $\bar r_{target} - \bar r_{initial}$ is the model's low-dimensional measurement that we are trying to reconstruct using the high-dimensional but sparse underlying signal, $\delta$.
This is  exactly the classical compressed sensing problem formulation.
As a result, we will add $l_1$ regularization to $\delta$ as an additional component to the loss function as is done in both linear and non-linear compressed sensing \cite{blumensath2013compressed}.  

Consequently, we could consider finding a \abr by minimizing the linear compressed sensing loss:
\begin{align}
loss_{cs}(\delta) = ||A(\bar x_{initial}  + \delta) - \bar r_{target}||_2^2 + \lambda ||\delta||_1 \label{eq:base}
\end{align}
where $\bar x_{i} = \frac{1}{|X_{i}|} \sum\limits_{x \in X_{i}} x$. 
 Or, by  removing the assumption that $r(x) = Ax$, minimizing the more general version that we use for our experiments:
\begin{equation} \label{eq:final}
    loss(\delta) = ||r(\bar x_{initial} + \delta) - \bar r_{target}||_2^2  + \lambda ||\delta||_1
\end{equation}  

\subsection{Computing \abrs} \label{sec:alg}

We have thus far limited ourselves to explaining the differences between $l=2$ groups of points labeled as $X_0$ (``initial'') and $X_1$ (``target''),  and focused on learning an explanation $t_{0 \rightarrow 1}$ from $X_0$ to $X_1$.  
However, this is not a realistic setting because this labeling was arbitrary and because we usually have $l > 2$.  

Unfortunately, the naive solution of using compressed sensing to independently produce explanations for all O($l^2$) pairs of groups will fail to satisfy two desirable properties related to the internal consistency of the explanations.  
For instance, we would like $t_{0 \rightarrow 1}$ to agree with $t_{1 \rightarrow 0}$, a property we call \emph{symmetry}.  
Additionally, we would like $t_{0 \rightarrow 2}$ to agree with the combined explanations $t_{0 \rightarrow 1}$ and $t_{1 \rightarrow 2}$;  we call this \emph{transitivity}. 
Formally, \textit{symmetry} requires that $t_{i \rightarrow j} = t_{j \rightarrow i}^{-1}$ and \textit{transitivity} requires that $t_{i \rightarrow k} = t_{j \rightarrow k} \circ t_{i \rightarrow j}$ for any $j$. 

Our approach to finding a consistent (\ie symmetrical and transitive)  set of explanations is to enforce the consistency constraints by-design.
We do this by computing a set of explanations relative to a \textit{reference group}.
We  assume that $X_0$ is the reference group and find a set of basis explanations $t_1, \ldots, t_{l-1}$, where $t_i = t_{0 \rightarrow i}$.
We  then use this set of basis explanations  to construct the explanation between any pair of the groups of points.\footnote{%
	Importantly, the transitivity constraint also ensures that our choice of how we label the groups or which group is the reference group does not influence the optimal solution of our algorithm.}   
Algorithm \ref{alg:construct} describes how $t_{i \rightarrow j}$ can be constructed from $t_1, \ldots, t_{l-1}$.

\begin{algorithm}[H]
	\small
	\caption{\small \name: Calculating \abrs with a Reference Group.  
		Note that, because $\lambda$ is applied to all of the explanations, we cannot tune it to guarantee that each explanation is exactly $k$-sparse.}
	\label{alg:explain}
	\begin{algorithmic}
		\STATE \textbf{Input:} Model: $r$ \\
		\hspace{30pt}Group Means: $\bar x_i$ (feature space) and \\
		\hspace{90pt} $\bar r_i$ (representation space)\\
		\hspace{90pt} for $i = 0, \ldots, l-1$ \\
		\hspace{30pt}$l_1$ Regularization Weight: $\lambda$\\
		\hspace{30pt}Learning Rate: $\alpha$\\
		\STATE Initialize: $\delta_1, \ldots, \delta_{l-1}$ to vectors of 0
		\WHILE{not converged}
		\STATE Sample $i \ne j$ from  $\{0, \ldots, l-1\}$
		\STATE Construct $t_{i \rightarrow j}$ $(\delta_{i \rightarrow j})$ using Algorithm \ref{alg:construct}
		\STATE Calculate objective:  $v = loss(\delta_{i \rightarrow j})$  using Equation \ref{eq:final}
		\STATE Update the components of $\delta_{i \rightarrow j}$ using Algorithm \ref{alg:update}
		\ENDWHILE
		\STATE \textbf{Return:} $\delta_1, \ldots, \delta_{l-1}$
	\end{algorithmic}
\end{algorithm}
\vspace{-10pt}
\begin{algorithm}[H]
	\small
	\caption{\small How to construct any explanation between an arbitrary pair of groups, $t_{i \rightarrow j}$, using  the set of basis explanations relative to the reference group, $t_1, \ldots, t_{l-1}$}
	\label{alg:construct}
	\begin{algorithmic}
		\STATE \textbf{Input:} $i, j$
		\IF{$i == 1$}
		\STATE \textbf{Return:} $t_j$
		\ELSIF{$j == 1$}
		\STATE \textbf{Return:} $t_i^{-1}$
		\ELSE
		\STATE \textbf{Return:} $t_j \circ t_i^{-1}$
		\ENDIF	
	\end{algorithmic}	
\end{algorithm}
\vspace{-10pt}
\begin{algorithm}[H]
	\small
	\caption{\small How to update the basis explanations, $\delta_1, \ldots, \delta_{l-1}$, based on the performance of $\delta_{i \rightarrow j}$. 
		This splits the signal from the gradient between the basis explanations used to construct $\delta_{i \rightarrow j}$.
		Note that it does not maintain any fixed level of sparsity.}	
	\label{alg:update}
	\begin{algorithmic}
		\STATE \textbf{Input:}  $i, j$, $\alpha$ (learning rate), $\nabla v$ (gradient of the loss function)
		\IF{$i == 1$}
		\STATE $\delta_j = \delta_j - \alpha \nabla v$
		\ELSIF{$j == 1$}
		\STATE $\delta_i = \delta_i + \alpha \nabla v$
		\ELSE
		\STATE $\delta_j = \delta_j - 0.5 \alpha \nabla v$ 
		\STATE $\delta_i = \delta_i + 0.5 \alpha \nabla v$
		\ENDIF
	\end{algorithmic}	
\end{algorithm}

\textbf{Overview of \name.}
We now have all of the pieces necessary to actually compute a \abr:  a differentiable loss function to measure the quality of $t_{i \rightarrow j}$, $l_1$ regularization to help us find the simplest possible explanation between $X_i$ and $X_j$, and a problem setup to ensure that our explanations are consistent across $X_0, \ldots, X_{l-1}$.  
At a high level, \name  will proceed to sample random ``initial'' and ``target'' groups from the set of all groups, construct that explanation from the set of basis explanations  (Algorithm \ref{alg:construct}), and then use Equation \ref{eq:final} as a loss function to use to update the explanation using gradient descent (Algorithm \ref{alg:update}).  
Pseudo-code for this process is in Algorithm \ref{alg:explain}.\footnote{The pseudo-code leaves out some of the details of the optimization process such as how often we sample new ``initial'' and ``target'' groups and how convergence is defined.  For those details, see the code on GitHub.}
The main hyper-parameter that requires tuning is the strength of the $l_1$ regularization, $\lambda$.

\textbf{Why can we prioritize correctness in Equation \ref{eq:final}?}
In the previous subsection, we noted that the Equation \ref{eq:final} prioritizes correctness over coverage. 
Because Algorithm \ref{alg:explain} randomly chooses the ``initial'' and ``target'' groups many times, it updates the basis explanations based on both $cr(\delta_{i \rightarrow j})$ and $cr(\delta_{j \rightarrow i})$.  
When $t$ is a translation and the explanations are symmetrical,  we can see that $cr(\delta_{j \rightarrow i})$ is closely related to $cv(\delta_{i \rightarrow j})$.  
This is because they only differ in whether they add $\delta_{j \rightarrow i}$ to a point in $X_j$ or subtract $\delta_{j \rightarrow i}$ from a point in $X_i$ in their respective indicator functions.
Further, if we consider $r(x) = Ax$, then they are identical metrics (this is consistent with the example from Figure \ref{fig:3-bad}).  
Collectively, this means that Algorithm \ref{alg:explain} implicitly considers both $cr(\delta_{i \rightarrow j})$ and $cv(\delta_{i \rightarrow j})$ while computing the explanations.

\subsection{Controlling the Level of Sparsity}
Because neither \name nor \base is guaranteed to produce a $k$-sparse explanation, we will threshold each of the explanations to include only the $k$ most important features (\ie the $k$ features with the largest absolute value) for our experiments.  
This is done after they have been calculated but before their quality has been evaluated.  
Importantly, \name has a hyper-parameter, $\lambda$, which roughly controls the sparsity of its explanations; as a result, we will tune $\lambda$ to maximize correctness for each value of $k$.   

The fact that $\lambda$ is tuned for each value of $k$ raises an interesting question:  ``Does \name use a subset of the features from its $k_2$-sparse explanation for its $k_1$-sparse explanation when $k_1 < k_2$?''.  
Naturally, we would like for the answer to be ``yes'' because, for example, it does not seem like a desirable outcome if a 2-sparse explanation uses Features A and B but a 1-sparse explanation uses Feature C.
  
Suppose we have two explanations between Group $i$ and Group $j$:  $e_1$ which is $k_1$-sparse and $e_2$ which is $k_2$-sparse with $k_1 < k_2$.  
To address this question, we define the \textit{similarity} of $e_1$ and $e_2$ as:
\begin{equation}
similarity(e_1, e_2) = \frac{\sum |e_1[i]| \mathbbm{1}[e_2[i] \ne 0]}{||e_1||_1}
\end{equation}
This metric captures how much of $e_1$'s explanation uses features that were also chosen by $e_2$.\footnote{Note that this metric also includes the run to run variance of \name.}
So a score of 1 indicates that $e_1$ uses a subset of the features of $e_2$ and a score of 0 indicates that it uses entirely different features.  
Because \base does not solve a different optimization problem to achieve each level of sparsity, its similarity measure is always 1.  
When we run experiments with a list of sparsity levels $k_1, \ldots, k_m$, we will plot $similarity(e_1, e_2), \ldots, similarity(e_{m-1}, e_m)$ to measure how similar \name's explanations are as the level of sparsity increases.

\section{Experimental Results}

Our experimental results are divided into two sections.  
In the first section, we demonstrate that \name is better at explaining the model than \base is when we restrict the explanations to varying degrees of sparsity.  
In the second section, we move beyond assessing whether or not \name explains the model and demonstrate that it also appears to capture real signals in the data.  

\subsection{\name's Efficacy at Explaining the Model}

\begin{figure}[]
	\centering
	\begin{subfigure}[t]{0.48\linewidth}
		\centering
		\includegraphics[width = \linewidth]{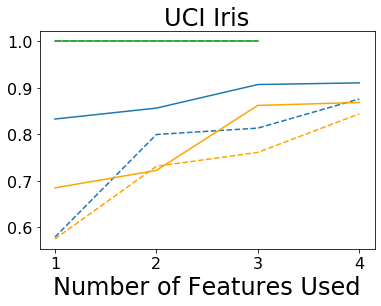}
	\end{subfigure}%
	\hspace{5pt}
	\begin{subfigure}[t]{0.48 \linewidth}
		\centering
		\includegraphics[width = \linewidth]{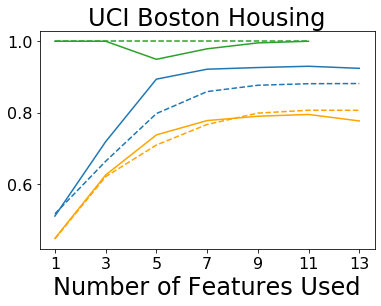}
	\end{subfigure}
	\begin{subfigure}[t]{0.48\linewidth}
		\centering
		\includegraphics[width = \linewidth]{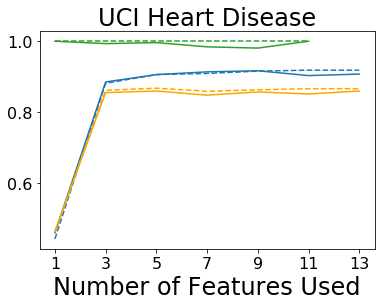}
	\end{subfigure}%
	\hspace{5pt}
	\begin{subfigure}[t]{0.48 \linewidth}
		\centering
		\includegraphics[width = \linewidth]{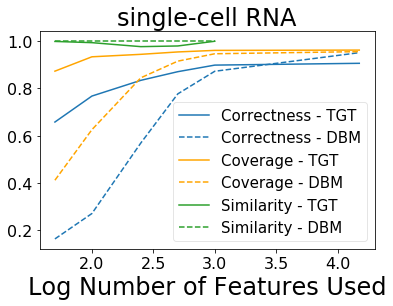}
	\end{subfigure}
	\caption{A comparison of the effectiveness of \name to \base at explaining the model (measured by correctness and coverage) at a range of sparsity levels.  
		Note that \name performs at least as well as \base and usually does better.  
		Looking at the similarity metric, we see that \name is fairly consistent at picking a subset of the current features when asked to find an even sparser solution.
	}
	\label{fig:tradeoff}
\end{figure}

From an interpretable machine learning perspective, our goal is help practitioners understand the dimensionality-reduction models they use in the data exploration process.
We measure the quality of \abrs  using correctness (Equation \ref{eq:cor}) and coverage (Equation \ref{eq:cov}) at varying degrees of sparsity (Figure \ref{fig:tradeoff}).

We use the model from \cite{ding2018interpretable}\footnote{%
We use this model because previous analysis showed that its representation identifies meaningful groups on the single-cell RNA dataset \cite{ding2018interpretable}.}%
on the UCI Iris, Boston Housing, and Heart Disease datasets \cite{Dua:2019} and a single-cell RNA dataset \cite{shekhar2016comprehensive} and use its a visualization of its two-dimensional representation to define the groups of points.
Figure \ref{fig:bipolar-rep} shows this representation and grouping for the single-cell RNA dataset;  similar plots for all of the datasets are in the Appendix \ref{appendix:rep}.  
This model learns a non-linear embedding using a neural network architecture which is trained to be a parametric version of t-SNE \cite{maaten2008visualizing} that also preserves the global structure in the data \cite{kobak2018art}.

Next, because the acceptable level of complexity depends on both the application and the person using the explanation, we measure the effectiveness of the explanations produced by \name and \base at explaining the model across a range of sparsity levels.  

\textbf{Explanation effectiveness at different levels of sparsity.}
Figure \ref{fig:tradeoff} shows the results of this comparison.  
We can see that \name performed at least as well as \base and usually did better.
Further, we can see that \name's explanations are quite similar to each other as we ask for sparser explanations.
Note that all of these metrics are defined for a single pair of groups and so these plots report the average across all pairs of groups.  

\begin{figure}[]
	\centering
	\begin{subfigure}[t]{0.48\linewidth}
		\centering
		\includegraphics[width = \linewidth]{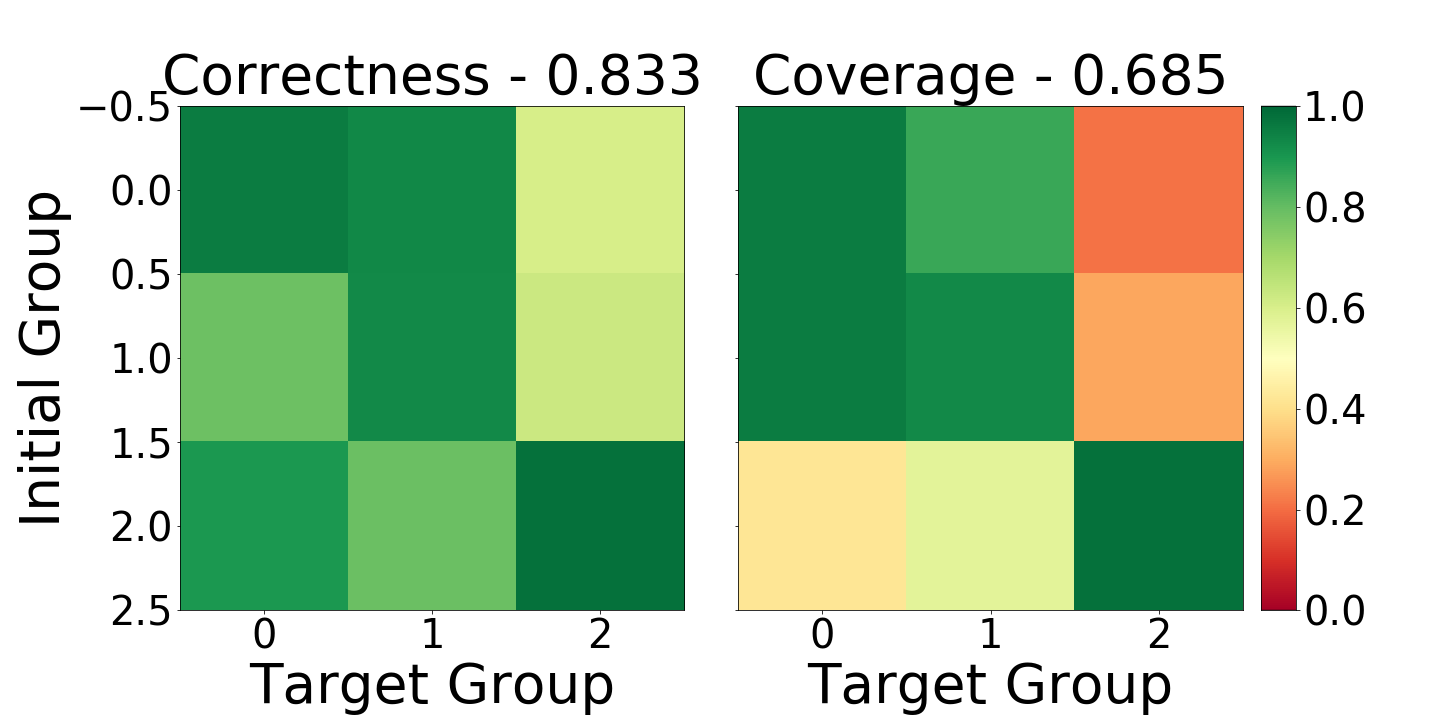}
	\end{subfigure}%
	\hspace{5pt}
	\begin{subfigure}[t]{0.48 \linewidth}
		\centering
		\includegraphics[width = \linewidth]{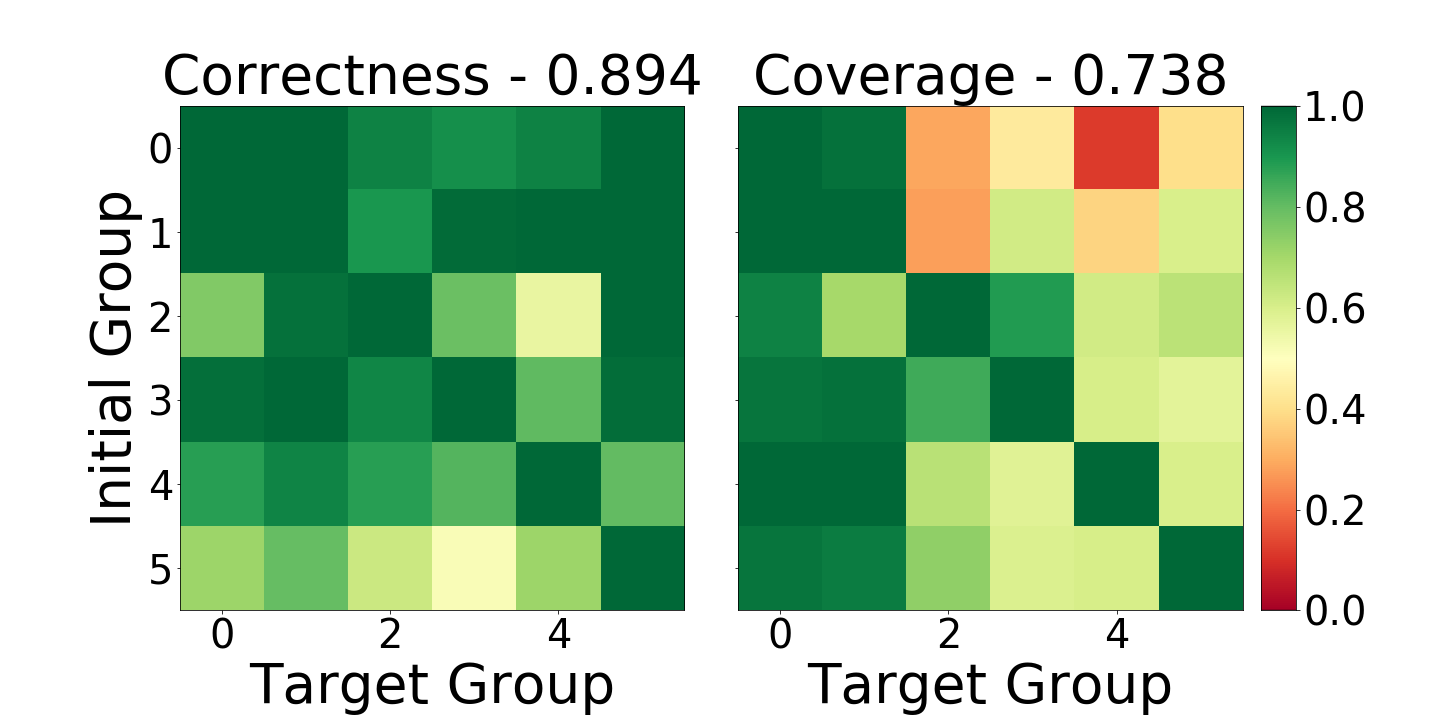}
	\end{subfigure}
	\begin{subfigure}[t]{0.48\linewidth}
		\centering
		\includegraphics[width = \linewidth]{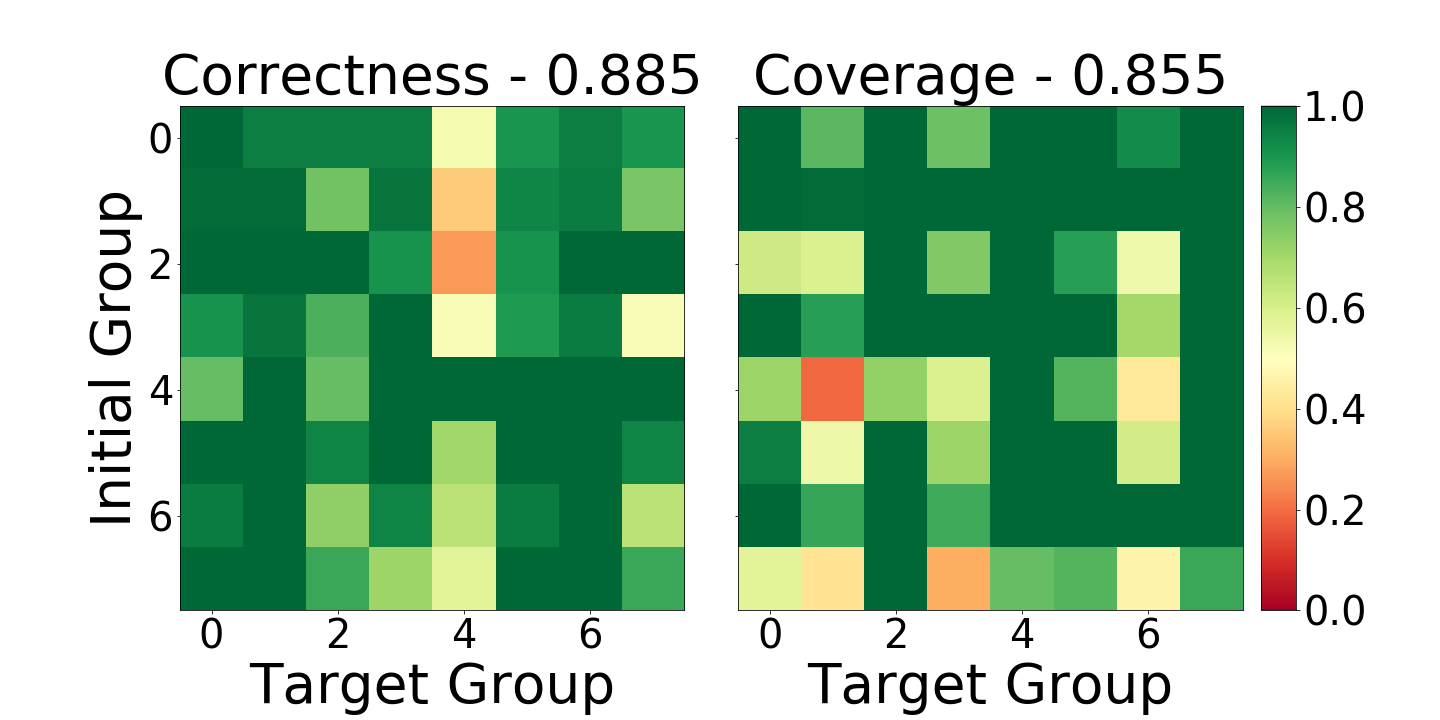}
	\end{subfigure}%
	\hspace{5pt}
	\begin{subfigure}[t]{0.48 \linewidth}
		\centering
		\includegraphics[width = \linewidth]{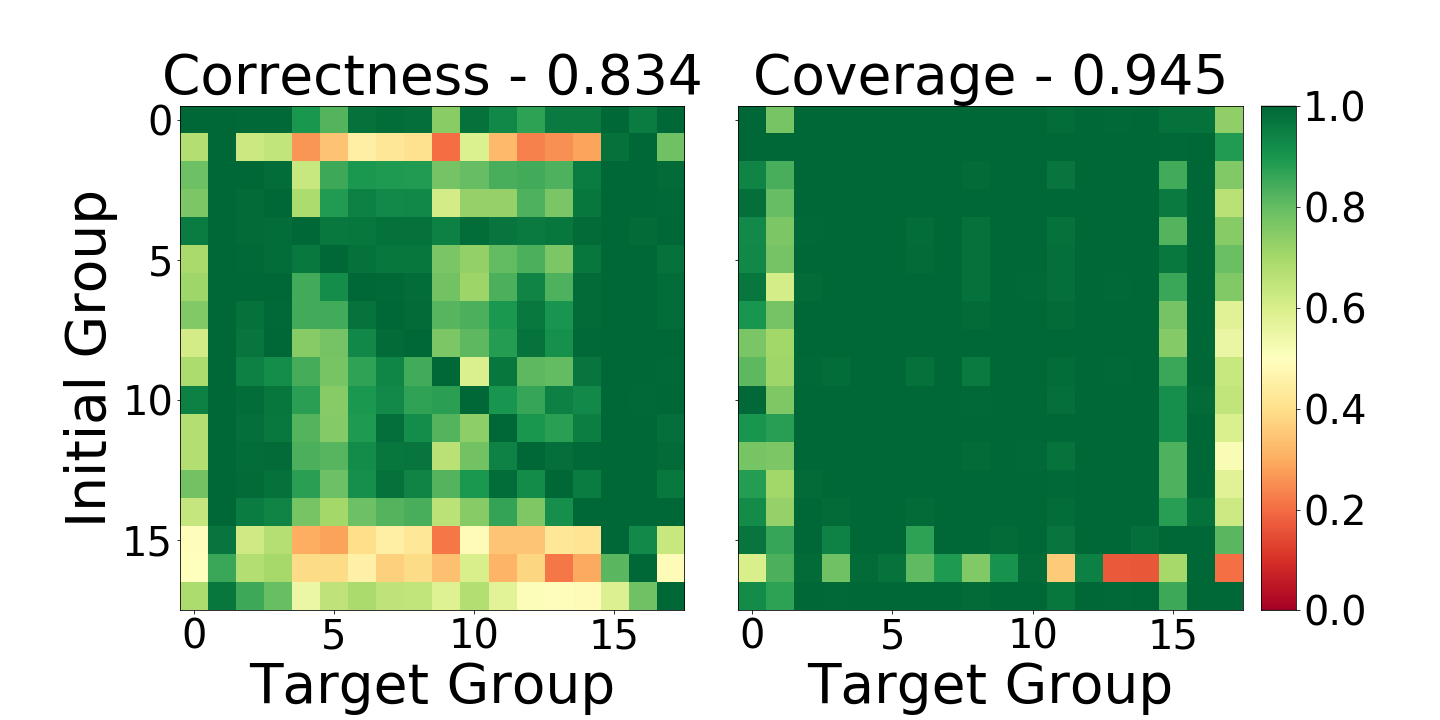}
	\end{subfigure}
	\caption{The pairwise explanation metrics for \name with:  Top Left) 1-sparse explanations on Iris, Top Right) 5-sparse explanations on Boston Housing, Bottom Left) 3-sparse explanations on Heart Disease, and Bottom Right) 250-sparse explanations on single-cell RNA.   
	}
	\label{fig:metrics}
	
	\includegraphics[width = 0.8\linewidth]{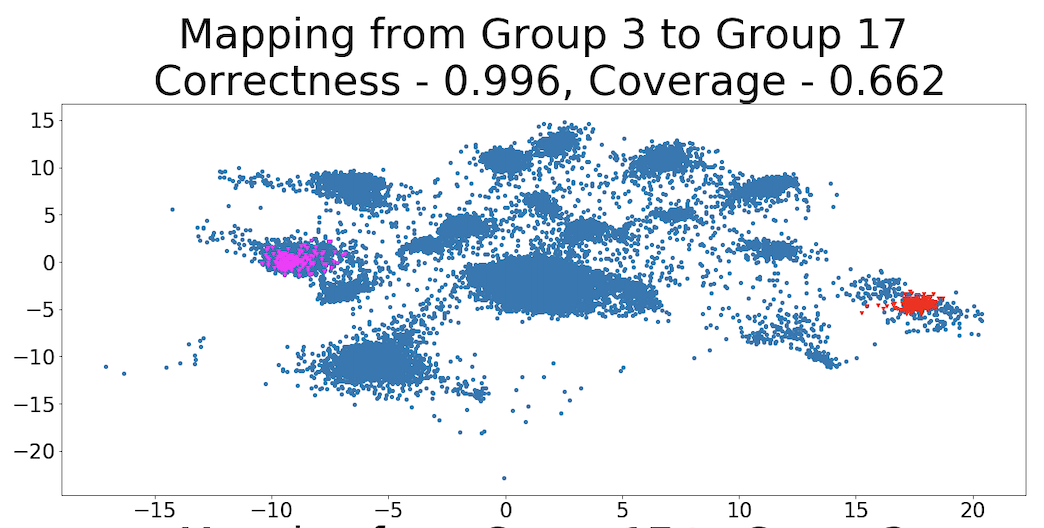}
	\captionof{figure}{Unlike \base, \name is able to produce an effective 250-sparse explanation for the difference between Group 3 and Group 17 on the single-cell RNA dataset.
	This can be seen both visually and with the correctness and coverage metrics.}
	\label{fig:tgt-example-applied}
\end{figure}

\textbf{Exploring a Specific Level of Sparsity.}
Figure \ref{fig:tradeoff} shows that \name's performance:  
 is almost as good when $k = 1$ as when $k = 4$ on Iris,
drops off sharply for $k < 5$ on Boston Housing,
and  drops off sharply for $k < 3$ on Heart Disease.
Further, on the single-cell RNA dataset, it shows that \name significantly outperforms \base when $k = 250$  (Appendix \ref{appendix:pairwise} Figure \ref{fig:a-bipolar-metrics-mean}) and that this comparison becomes more favorable for \name for smaller $k$.  
The level of sparsity where the metrics drop off indicates the minimum explanation complexity required for these methods to explain the model.  
See Figure \ref{fig:metrics} for an example of the pairwise correctness and coverage metrics for these levels of sparsity.

Figure \ref{fig:bipolar-dbm-example-applied} shows that \base does not produce a good 250-sparse explanation for the difference between Group 3 and Group 17 from Figure \ref{fig:bipolar-rep}.  
For the sake of an easy comparison, Figure \ref{fig:tgt-example-applied} shows a similar plot that uses \name's 250-sparse explanation;  it is clearly a much better explanation.  

\subsection{\name's Efficacy at Capturing Real Signals in the Data}

In the previous section, we demonstrated that \name provides accurate explanations of the model that learned the low-dimensional representation.  
However, in practice, there could be a mismatch between what the model itself learns and the true underlying structure in the data.
In this section, we evaluate empirically whether or not \name provides explanations that match underlying patterns in the data.

We begin with an experiment on a synthetic dataset with a known causal structure and demonstrate that \name correctly identifies this structure.  
This also serves as an intuitive example of why a sparser explanation can be as effective as a less-sparse explanation.  
Next, we leverage the labels that come with the UCI datasets to compare \name's explanations to some basic domain knowledge.  
Finally, we modify the UCI datasets and demonstrate that \name is able to identify those modifications.  
Together, these results indicate that \name is identifying real patterns in the data.  

\textbf{Synthetic Data with a Known Causal Structure.}
By specifying the causal structure of the data, we can know which differences are necessary in the explanation, since they are the casual differences, and which differences are unnecessary, since they are explained by the causal differences.  
We find that \name correctly identifies the causal structure of this dataset and that \base does not.  

We use the following procedure to generate each point in our synthetic dataset: $x_1, x_2 \sim Bern(0.5) + \mathcal{N}(0, 0.2)$, $x_3 \sim \mathcal{N}(0, 0.5)$, and $x_4 \sim x_1 + \mathcal{N}(0, 0.05)$.  
The causal structure of this dataset is simple.  
\textit{$x_1$ and $x_2$ jointly cause 4 different groups of points.}  
The explanation for the differences between these groups must include these variables.  
\textit{$x_3$ is a noise variable} that is unrelated to these groups and, as a result, should not be included in any explanation.  
\textit{$x_4$ is a variable that is correlated with those groups}, since it is caused by $x_1$, but does not cause those groups.  
As a result, it is not necessary to include it in any explanation.

We generate a dataset consisting of 400 points created using this process and train an autoencoder \cite{kramer1991nonlinear} to learn a two dimensional representation of the dataset. 
A visualization of this learned representation is in the Appendix \ref{appendix:rep} Figure \ref{fig:a-synth-rep}; as expected, there are four distinct groups of points in it. 
Then, we use \name and \base to calculate the \abrs between these groups.  
The pairwise and average correctness and coverage metrics for these solutions are in the Appendix \ref{appendix:pairwise} Figures \ref{fig:a-synth-metrics}  and \ref{fig:a-synth-metrics-mean};  observe that the two methods are equally effective at explaining the model.  

When we inspect the explanations ()Table \ref{tab:synth}), we see that both \name and \base use $x_1$ and $x_2$, neither use $x_3$, and that \base uses $x_4$ while \name does not.  
This shows that, even in a very simple setting, there is good reason to believe that an explanation that is simpler (\ie sparser) than the \base explanation exists and that \name might be able to find it.  

\begin{table}[]
	\centering
	\captionof{table}{A comparison of the explanations from \name and from \base.  
		Note that they are similar except that \name does not use $x_4$, which is the variable that is not causaly related to the groups.}
	\resizebox{\linewidth}{!}{
		\begin{tabular}{@{}llllll@{}}
			\toprule
			Explanation & Method & $x_1$ & $x_2$ & $x_3$ & $x_4$ \\ \midrule
			$0 \rightarrow 1$ & \name & -1.09 & 0.01 & 0.03 & 0.00 \\
			& \base & -1.02 & 0.04 & 0.01 & -1.03 \\
			$0 \rightarrow 2$ & \name & 0.00 & 0.88 & 0.00 & 0.00 \\
			& \base & -0.01 & 0.97 & 0.06 & -0.03 \\
			$0 \rightarrow 3$ & \name & -0.99 & 0.71 & 0.00 & 0.00 \\
			& \base & -1.02 & 1.03 & -0.01 & -1.03 \\ \bottomrule
	\end{tabular}}
	\label{tab:synth}
\end{table}

\textbf{Qualitative Analysis of the UCI Datasets using the Labels.}
Qualitatively, we find that \name's explanations agree with domain knowledge about these datasets.  
Specifically:  
On the Iris dataset, its explanations agree with a simple decision tree because they both rely mostly on the Petal Width to separate the groups.  
On the Boston Housing dataset, it identifies the differences between a set of inexpensive urban houses vs expensive suburban houses as well as equally priced groups of houses that differ mainly in whether or not they are on the Charles River.  
Finally on the Heart Disease dataset, it finds that the difference between a moderate and a low risk group of subjects was that the low-risk group's symptoms are explained by something other than heart disease and the difference between the moderate and high risk group of subjects is that the former is made up of men and the later of women.  
For full details, see the Appendix \ref{appendix:uci}.  

\textbf{Quantitative Analysis of Modified Versions of the UCI Datasets.}
In order to perform a more quantitative analysis, we artificially add a known signal to the dataset by choosing one of the groups of points, creating a modified copy of it by translating it, and adding those new points back into the dataset.  
We then ask two important questions about \name's behavior.  
First, does \name correctly identify the modifications we made to the original dataset?
Second, do \name's explanations between the original groups change when the modified group is added to the dataset? 
Details of how we setup these experiments and their results are in the Appendix \ref{appendix:modify}.  

We find that \name does identify the modifications we made and that, in doing so, it does not significantly change the explanations between the original groups. 
Importantly, this result remains true even if we retrain the learned representation on the modified dataset.  

These results are a strong indicator that \name finds real patterns in the data because it recovers both the original signal and the artificial signal even when the algorithm is rerun or the representation is retrained.

\section{Conclusion}

In this work, we introduced a new type of  explanation, a \abr, which is a counterfactual explanation that applies to an entire group of points rather than a single point.  
Next, we defined reasonable metrics to measure the quality of \abrs (\ie correctness and coverage) and introduced the concept of consistency (\ie symmetry and transitivity), which is an important additional criteria that \abrs must satisfy.  
Given that, we defined an algorithm for finding consistent \abrs, \name, that treats each pairwise explanation as a compressed sensing problem.  
Our first experiments empirically demonstrated that \name is better able to explain the model than \base across a range of levels of sparsity.  
Our next experiments showed that \name captures real patterns in the data.
This was done using a synthetic dataset with a known causal structure and by comparing \name's explanations to background knowledge about the UCI datasets.  
As an additional test, we then added a synthetic signal to the UCI datasets and demonstrated that \name can recover that signal without changing its explanations between the real groups.  
Importantly, this result remains true even when the representation is retrained.  

Although we focused on data exploration in this work, similar groups arise naturally whenever the model being trained uses an encoder-decoder structure.  
This technique is ubiquitous in most areas where deep learning is common (\eg semi-supervised learning, image classification, natural language processing, reinforcement learning).  
In these settings, identifying the key differences between the groups is an interesting question because we can observe that ``The model treats points in Group A the same/differently as points in Group B'', determine that ``The key differences between Group A and Group B are X'', and then conclude that ``The model does not/does use pattern X to make its decisions''.  
We believe that exploring these applications is an important direction of future work that will require more sophisticated transformation functions, optimization procedures, and definitions of the groups of points of interest.   

\section*{Acknowledgments}
This work was supported in part by DARPA FA875017C0141, the National Science Foundation grants IIS1705121 and IIS1838017, an Okawa Grant, a Google Faculty Award, an Amazon Web Services Award, a JP Morgan A.I. Research Faculty Award, and a Carnegie Bosch Institute Research Award. Any opinions, findings and conclusions or recommendations expressed in this material are those of the author(s) and do not necessarily reflect the views of DARPA, the National Science Foundation, or any other funding agency.
We would also like to thank Joon Kim, Jeffrey Li, Valerie Chen, Misha Khodak, and Jeremy Cohen for their feedback while writing this paper.  

\bibliography{bib.bib}

\begin{thebibliography}{28}
\providecommand{\natexlab}[1]{#1}
\providecommand{\url}[1]{\texttt{#1}}
\expandafter\ifx\csname urlstyle\endcsname\relax
  \providecommand{\doi}[1]{doi: #1}\else
  \providecommand{\doi}{doi: \begingroup \urlstyle{rm}\Url}\fi

\bibitem[Bello \& Mosca(2004)Bello and Mosca]{bello2004epidemiology}
Bello, N. and Mosca, L.
\newblock Epidemiology of coronary heart disease in women.
\newblock \emph{Progress in cardiovascular diseases}, 46\penalty0 (4):\penalty0
  287--295, 2004.

\bibitem[Blumensath(2013)]{blumensath2013compressed}
Blumensath, T.
\newblock Compressed sensing with nonlinear observations and related nonlinear
  optimization problems.
\newblock \emph{IEEE Transactions on Information Theory}, 59\penalty0
  (6):\penalty0 3466--3474, 2013.

\bibitem[Cand{\`e}s et~al.(2006)]{candes2006compressive}
Cand{\`e}s, E.~J. et~al.
\newblock Compressive sampling.
\newblock In \emph{Proceedings of the international congress of
  mathematicians}, volume~3, pp.\  1433--1452. Madrid, Spain, 2006.

\bibitem[Caruana et~al.(2015)Caruana, Lou, Gehrke, Koch, Sturm, and
  Elhadad]{caruana2015intelligible}
Caruana, R., Lou, Y., Gehrke, J., Koch, P., Sturm, M., and Elhadad, N.
\newblock Intelligible models for healthcare: Predicting pneumonia risk and
  hospital 30-day readmission.
\newblock In \emph{Proceedings of the 21th ACM SIGKDD International Conference
  on Knowledge Discovery and Data Mining}, pp.\  1721--1730. ACM, 2015.

\bibitem[Dhurandhar et~al.(2018)Dhurandhar, Chen, Luss, Tu, Ting, Shanmugam,
  and Das]{dhurandhar2018explanations}
Dhurandhar, A., Chen, P.-Y., Luss, R., Tu, C.-C., Ting, P., Shanmugam, K., and
  Das, P.
\newblock Explanations based on the missing: Towards contrastive explanations
  with pertinent negatives.
\newblock In \emph{Advances in Neural Information Processing Systems}, pp.\
  592--603, 2018.

\bibitem[Dhurandhar et~al.(2019)Dhurandhar, Pedapati, Balakrishnan, Chen,
  Shanmugam, and Puri]{dhur2019model}
Dhurandhar, A., Pedapati, T., Balakrishnan, A., Chen, P.-Y., Shanmugam, K., and
  Puri, R.
\newblock Model agnostic contrastive explanations for structured data, 2019.

\bibitem[Ding et~al.(2018)Ding, Condon, and Shah]{ding2018interpretable}
Ding, J., Condon, A., and Shah, S.~P.
\newblock Interpretable dimensionality reduction of single cell transcriptome
  data with deep generative models.
\newblock \emph{Nature communications}, 9\penalty0 (1):\penalty0 2002, 2018.

\bibitem[Dua \& Graff(2017)Dua and Graff]{Dua:2019}
Dua, D. and Graff, C.
\newblock {UCI} machine learning repository, 2017.
\newblock URL \url{http://archive.ics.uci.edu/ml}.

\bibitem[Goyal et~al.(2019)Goyal, Wu, Ernst, Batra, Parikh, and
  Lee]{goyal2019counterfactual}
Goyal, Y., Wu, Z., Ernst, J., Batra, D., Parikh, D., and Lee, S.
\newblock Counterfactual visual explanations.
\newblock In \emph{International Conference on Machine Learning}, pp.\
  2376--2384, 2019.

\bibitem[Jiang et~al.(2004)Jiang, Tang, and Zhang]{jiang2004cluster}
Jiang, D., Tang, C., and Zhang, A.
\newblock Cluster analysis for gene expression data: a survey.
\newblock \emph{IEEE Transactions on knowledge and data engineering},
  16\penalty0 (11):\penalty0 1370--1386, 2004.

\bibitem[Kauffmann et~al.(2019)Kauffmann, Esders, Montavon, Samek, and
  M{\"u}ller]{kauffmann2019clustering}
Kauffmann, J., Esders, M., Montavon, G., Samek, W., and M{\"u}ller, K.-R.
\newblock From clustering to cluster explanations via neural networks.
\newblock \emph{arXiv preprint arXiv:1906.07633}, 2019.

\bibitem[Kobak \& Berens(2018)Kobak and Berens]{kobak2018art}
Kobak, D. and Berens, P.
\newblock The art of using t-sne for single-cell transcriptomics.
\newblock \emph{bioRxiv}, pp.\  453449, 2018.

\bibitem[Kramer(1991)]{kramer1991nonlinear}
Kramer, M.~A.
\newblock Nonlinear principal component analysis using autoassociative neural
  networks.
\newblock \emph{AIChE journal}, 37\penalty0 (2):\penalty0 233--243, 1991.

\bibitem[Lipton(2016)]{lipton2016mythos}
Lipton, Z.~C.
\newblock The mythos of model interpretability.
\newblock \emph{arXiv preprint arXiv:1606.03490}, 2016.

\bibitem[Lundberg \& Lee(2017)Lundberg and Lee]{lundberg2017unified}
Lundberg, S.~M. and Lee, S.-I.
\newblock A unified approach to interpreting model predictions.
\newblock In \emph{Advances in Neural Information Processing Systems}, pp.\
  4765--4774, 2017.

\bibitem[Maaten \& Hinton(2008)Maaten and Hinton]{maaten2008visualizing}
Maaten, L. v.~d. and Hinton, G.
\newblock Visualizing data using t-sne.
\newblock \emph{Journal of machine learning research}, 9\penalty0
  (Nov):\penalty0 2579--2605, 2008.

\bibitem[Plumb et~al.(2018)Plumb, Molitor, and Talwalkar]{plumb2018model}
Plumb, G., Molitor, D., and Talwalkar, A.~S.
\newblock Model agnostic supervised local explanations.
\newblock In \emph{Advances in Neural Information Processing Systems}, pp.\
  2515--2524, 2018.

\bibitem[Ribeiro et~al.(2016)Ribeiro, Singh, and Guestrin]{ribeiro2016should}
Ribeiro, M.~T., Singh, S., and Guestrin, C.
\newblock Why should i trust you?: Explaining the predictions of any
  classifier.
\newblock In \emph{Proceedings of the 22nd ACM SIGKDD International Conference
  on Knowledge Discovery and Data Mining}, pp.\  1135--1144. ACM, 2016.

\bibitem[Ribeiro et~al.(2018)Ribeiro, Singh, and Guestrin]{ribeiro2018anchors}
Ribeiro, M.~T., Singh, S., and Guestrin, C.
\newblock Anchors: High-precision model-agnostic explanations.
\newblock AAAI, 2018.

\bibitem[Shaffer(1995)]{shaffer1995multiple}
Shaffer, J.~P.
\newblock Multiple hypothesis testing.
\newblock \emph{Annual review of psychology}, 46\penalty0 (1):\penalty0
  561--584, 1995.

\bibitem[Shekhar et~al.(2016)Shekhar, Lapan, Whitney, Tran, Macosko, Kowalczyk,
  Adiconis, Levin, Nemesh, Goldman, et~al.]{shekhar2016comprehensive}
Shekhar, K., Lapan, S.~W., Whitney, I.~E., Tran, N.~M., Macosko, E.~Z.,
  Kowalczyk, M., Adiconis, X., Levin, J.~Z., Nemesh, J., Goldman, M., et~al.
\newblock Comprehensive classification of retinal bipolar neurons by
  single-cell transcriptomics.
\newblock \emph{Cell}, 166\penalty0 (5):\penalty0 1308--1323, 2016.

\bibitem[Spall(1998)]{spall1998overview}
Spall, J.~C.
\newblock An overview of the simultaneous perturbation method for efficient
  optimization.
\newblock \emph{Johns Hopkins apl technical digest}, 19\penalty0 (4):\penalty0
  482--492, 1998.

\bibitem[Sundararajan et~al.(2017)Sundararajan, Taly, and
  Yan]{sundararajan2017axiomatic}
Sundararajan, M., Taly, A., and Yan, Q.
\newblock Axiomatic attribution for deep networks.
\newblock In \emph{Proceedings of the 34th International Conference on Machine
  Learning-Volume 70}, pp.\  3319--3328. JMLR. org, 2017.

\bibitem[Szegedy et~al.(2013)Szegedy, Zaremba, Sutskever, Bruna, Erhan,
  Goodfellow, and Fergus]{szegedy2013intriguing}
Szegedy, C., Zaremba, W., Sutskever, I., Bruna, J., Erhan, D., Goodfellow, I.,
  and Fergus, R.
\newblock Intriguing properties of neural networks.
\newblock \emph{arXiv preprint arXiv:1312.6199}, 2013.

\bibitem[Tomsett et~al.(2019)Tomsett, Harborne, Chakraborty, Gurram, and
  Preece]{tomsett2019sanity}
Tomsett, R., Harborne, D., Chakraborty, S., Gurram, P., and Preece, A.
\newblock Sanity checks for saliency metrics, 2019.

\bibitem[Tsaig \& Donoho(2006)Tsaig and Donoho]{tsaig2006extensions}
Tsaig, Y. and Donoho, D.~L.
\newblock Extensions of compressed sensing.
\newblock \emph{Signal processing}, 86\penalty0 (3):\penalty0 549--571, 2006.

\bibitem[Wang \& Rudin(2015)Wang and Rudin]{wang2015falling}
Wang, F. and Rudin, C.
\newblock Falling rule lists.
\newblock In \emph{Artificial Intelligence and Statistics}, pp.\  1013--1022,
  2015.

\bibitem[Zhang et~al.(2018)Zhang, Solar-Lezama, and
  Singh]{zhang2018interpreting}
Zhang, X., Solar-Lezama, A., and Singh, R.
\newblock Interpreting neural network judgments via minimal, stable, and
  symbolic corrections.
\newblock In \emph{Advances in Neural Information Processing Systems}, pp.\
  4874--4885, 2018.

\end{thebibliography}
\bibliographystyle{icml2020}

\clearpage

\appendix
\section{Appendix}

\subsection{An Example of the Difficulty of Using Existing Methods}
\label{appendix:rw}

For this example, we are going to consider Integrated Gradients (IG) \cite{sundararajan2017axiomatic} which produces local feature attribution explanations. 
Because IG is a supervised method, we start by training a classifier on top of the learned representation to get a multi-class classification model $f$ that predicts which group a point belongs to.  
Because our goal is to explain the difference between Group A and Group B with IG, we average IG's explanation for each point in Group B relative to each possible baseline value of a point in Group A for $f$'s Class B label.  
To be more precise:
\begin{equation}
\tiny
\label{eq:gig}
\delta_{IG}(A \rightarrow B) = \frac{1}{|X_A|\ |X_B|} \sum\limits_{x \in X_B} \sum\limits_{a \in X_A} IG(x, \text{class} = B, \text{baseline} = a)
\end{equation}
We will refer to this as `group Integrated Gradients' or gIG.  

\textbf{Challenge 1: Comparing Explanation Types.}
Because IG produces feature attributions and \name produces counterfactuals, there is no reliable metric in the literature to directly compare them.  
On the one hand, most feature attributions are the `correct explanation' for their specific definitions for `attribution' and the `baseline' value; this has made measuring their quality challenging \cite{tomsett2019sanity}.  
On the other hand, we cannot treat a feature attribution as a transformation function/translation, so our metrics and other metrics for counterfactual explanations cannot be applied.  

As a result, we compare \name to gIG on the same synthetic dataset we used earlier.  
We found that gIG identifies the causal variables as being significant and ignores the noise variable, but that it also identifies the correlated variable as being significant.  
This indicates that it is likely to be unable to find sparse explanations as well as \name can.  

\textbf{Challenge 2:  Consistency of Aggregated Local Explanations.}
One of the reasons we chose IG as a baseline method to aggregate is because its attributions are symmetrical and transitive \emph{with respect to a fixed class}. 
In other words, if all we cared about was explaining the differences between all of the groups of points with respect to a single reference group, say Group C, then IG would produce consistent explanations.  
However, explaining the features that separate Group A from Group B relative to Group C is not the problem we are trying to solve.   

When we use Equation \ref{eq:gig} to calculate $\delta_{IG}(i \rightarrow j)$ we found that the resulting explanations were not consistent.  
This does not violate the theory of IG because each $\delta_{IG}(i \rightarrow j)$ is calculated with reference Group $j$ and so the assumption that we have a single reference group is not satisfied.  

When we considered modifying Equation \ref{eq:gig} to aggregate over the reference `class/group' and potentially gain consistency that way, we either got uniform zero attributions (if we averaged over all reference groups) or inconsistent explanations (if we excluded any subset of $\{i, j\}$ from the averaging).  
 
\textbf{Conclusion.}
As suggested in Section \ref{sec:RelatedWork}, using existing explanation methods to find \abrs is going to be challenging because it is not what they are designed to do.  
We found that IG, a method that theoretically looked promising, was unable to be extended in a simple way to this setting.

\subsection{Representation Function}
\label{appendix:function}

\textbf{Differentiability.}
\name assumes that $r$ is a differentiable function.  
Hidden in this assumption is the assumption that $r$ is a function that we can evaluate on an arbitrary point.  
Although most methods for learning a low-dimensional representation satisfy this assumption, t-SNE does not.  
Fortunately there are parametric variations of t-SNE such as the one we used in our experiments \cite{ding2018interpretable}.  
The assumption that $r$ is differentiable can be relaxed by using a finite-difference optimization method, such as SPSA \cite{spall1998overview}, at the expense of computational cost.  

\textbf{Learning Meaningful Structure.}
One assumption that every analysis (whether that is manual inspection, statistical testing, or interpretable ML) of the representation learned by $r$ is that this function learned meaningful structure from the data.
Because practitioners are already relying on these representations and, in some situations, have verified that they are meaningful, this concern is largely orthogonal to our work.  

However, from an interpretable ML perspective, our goal is to explain $r$.  
So, if $r$ identifies different structure when it is retrained or when it is trained with a different algorithm or structure, we expect \name to produce different explanations since the embedding itself has changed.  

Our experimental results show that the representation learned by \cite{ding2018interpretable} is stable to being retrained and to modifications to the dataset and that \name produces stable explanations for these representations.  

\textbf{Identifying that Structure with Explanations.}
It is possible to have a model that learned the true structure of the data and to have an explanation that is technically true (as measured by some proxy metric for interpretability) about the model but that also fails to capture meaningful patterns.  
For example, adversarial examples \cite{szegedy2013intriguing} are technically local counterfactual explanations but they usually look like random noise and, as a result, do not tell a person much about the patterns the model has learned.  
\name's design, which calculates the explanation between each pair of groups as if it were a compressed sensing problem but constrains those solutions to be symmetrical and transitive among all groups, was chosen as a prior to prevent this type of behavior. 

\subsection{Learned Representations}
\label{appendix:rep}

\begin{figure}[h]
	\centering
	\includegraphics[width = 0.5\linewidth]{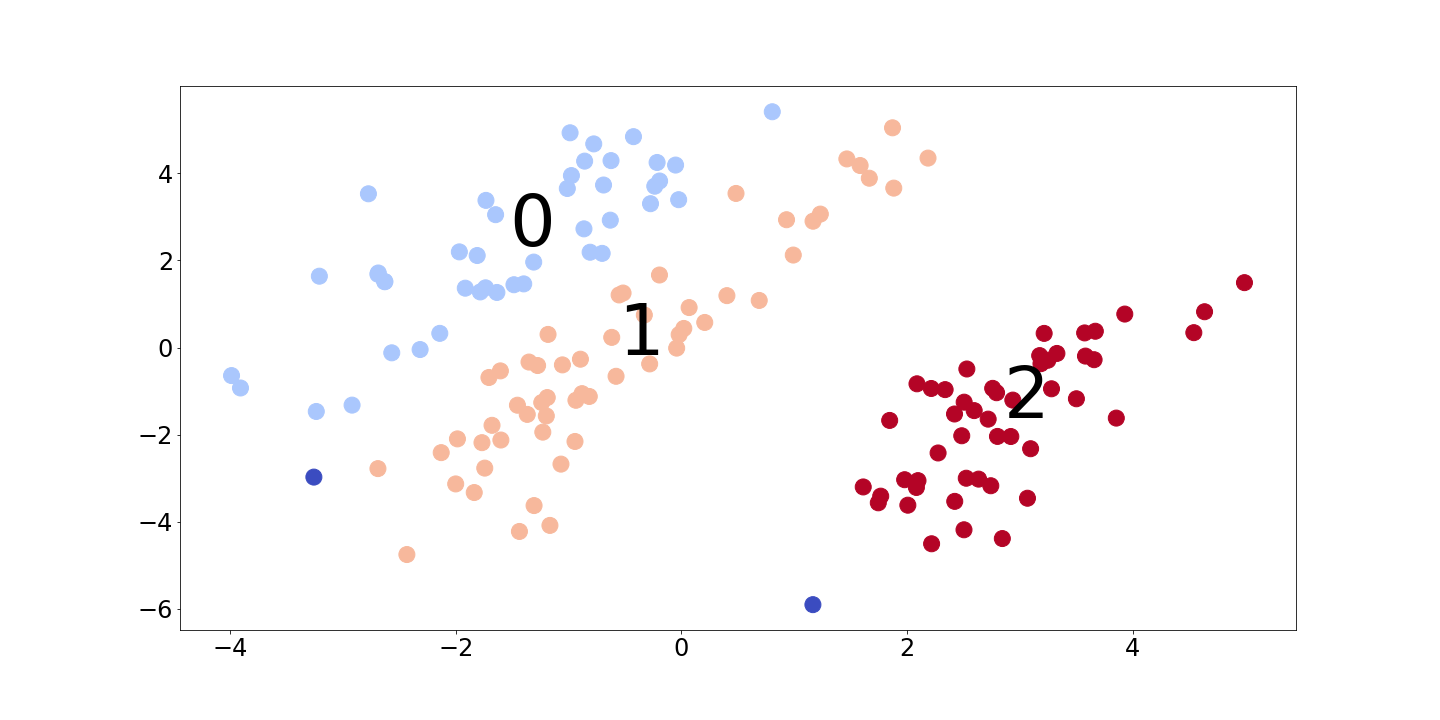}
	\caption{The learned representation and grouping for the UCI Iris dataset}
	\label{fig:a-iris-rep}
	\includegraphics[width = 0.5\linewidth]{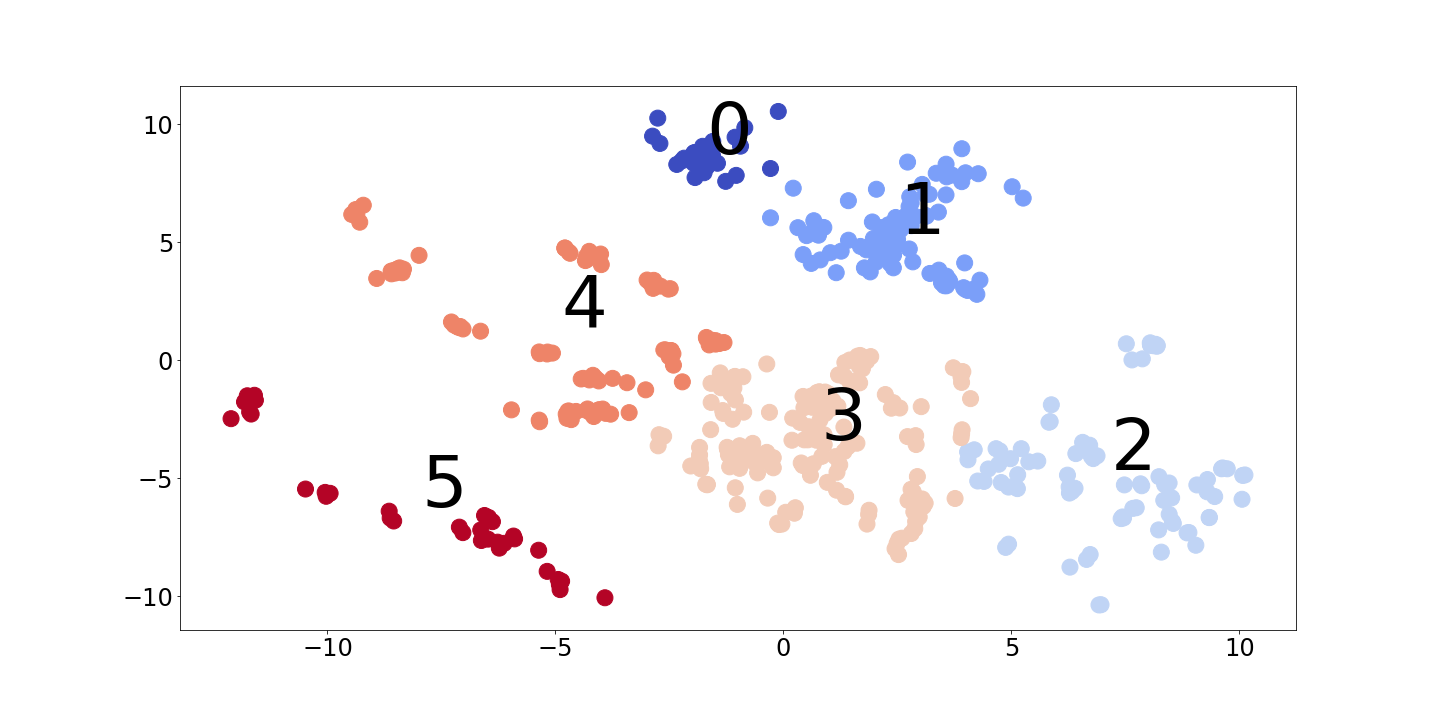}
	\caption{The learned representation and grouping for the UCI Boston Housing dataset}
	\label{fig:a-housing-rep}
	\includegraphics[width = 0.5\linewidth]{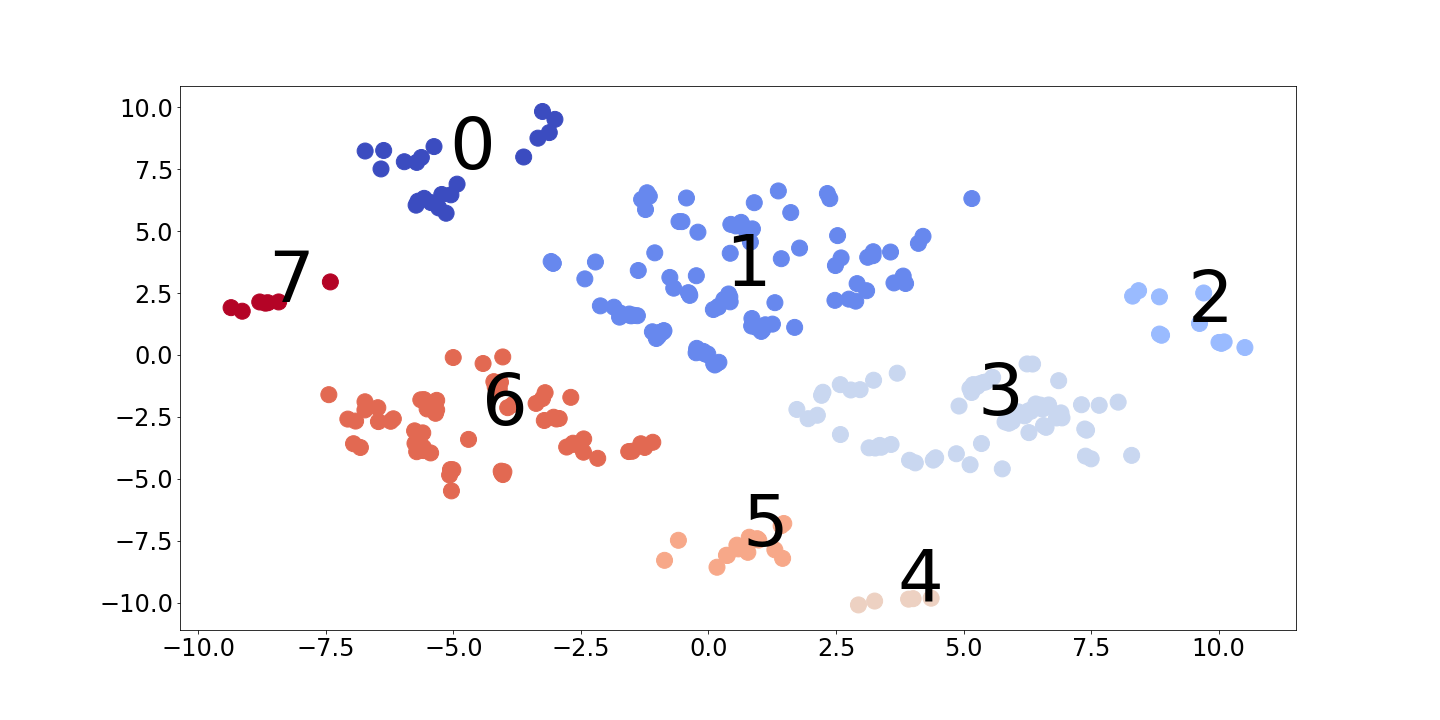}
	\caption{The learned representation and grouping for the UCI Heart Disease dataset}
	\label{fig:a-heart-rep}
	\includegraphics[width = 0.5\linewidth]{Figures/representations/bipolar-rep.png}
	\caption{The learned representation and grouping for the single-cell RNA dataset}
	\label{fig:a-bipolar-rep}
	\includegraphics[width = 0.5\linewidth]{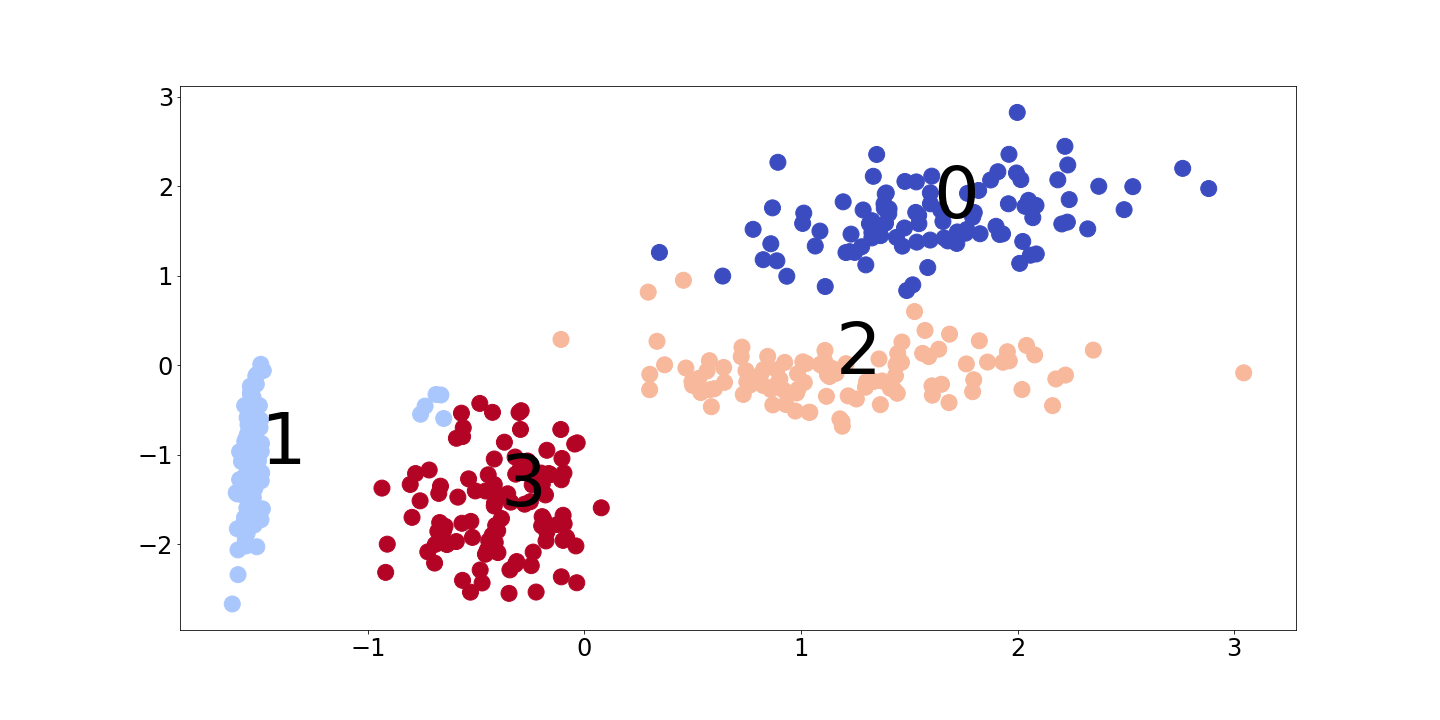}
	\captionof{figure}{The learned representation and grouping for the synthetic dataset.}
	\label{fig:a-synth-rep}
\end{figure}

The learned representations and the corresponding groups of points for the datasets we studied are in Figures \ref{fig:a-iris-rep}, \ref{fig:a-housing-rep}, \ref{fig:a-heart-rep},   \ref{fig:a-bipolar-rep}, and \ref{fig:a-synth-rep}.  

\subsection{Pairwise Correctness and Coverage Plots}
\label{appendix:pairwise}

\begin{figure}[h]
	\centering
	\includegraphics[width = 0.5\linewidth]{Figures/metrics/bipolar-metrics.png}
	\caption{The pairwise metrics for \name on the single-cell RNA dataset for 250-sparse explanations.}
	\label{fig:a-bipolar-metrics}
	\includegraphics[width = 0.5\linewidth]{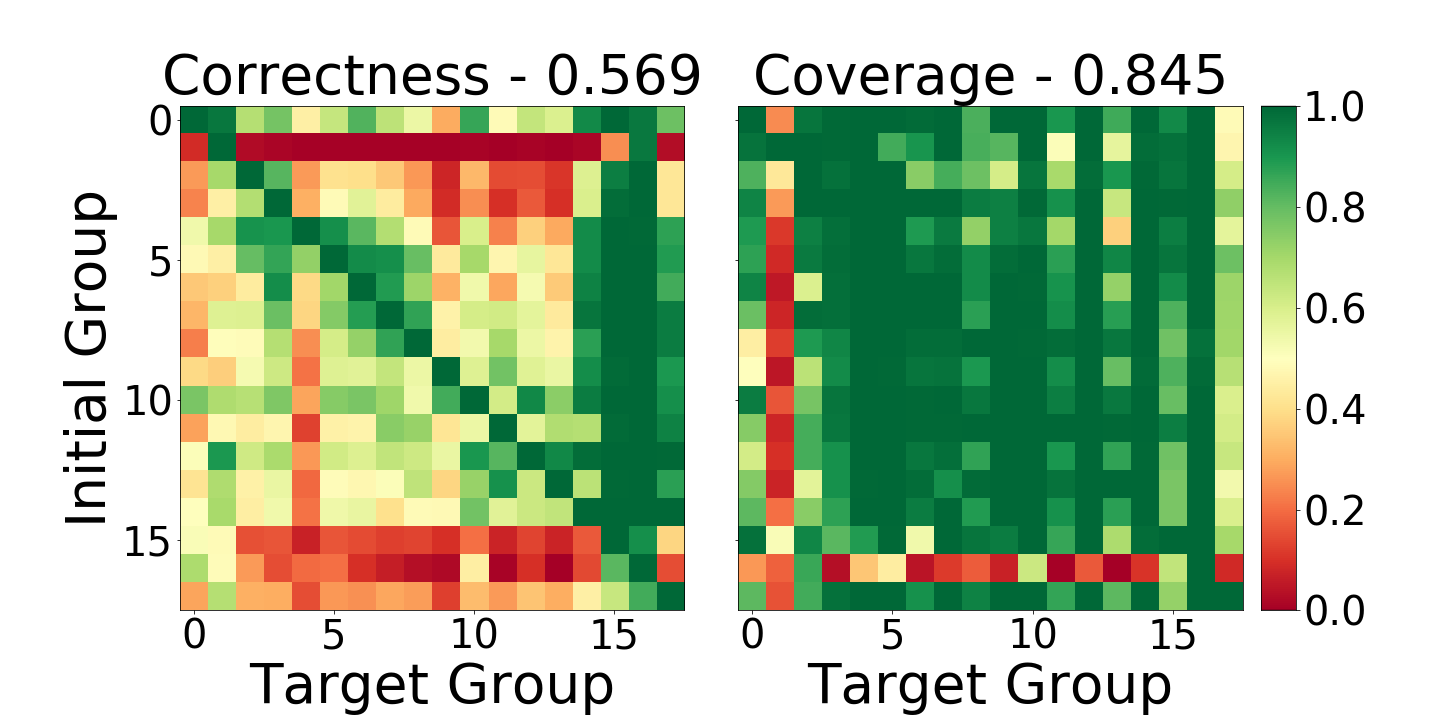}
	\caption{The pairwise metrics for \base on the single-cell RNA dataset for 250-sparse explanations.}
	\label{fig:a-bipolar-metrics-mean}
	\includegraphics[width = 0.5\linewidth]{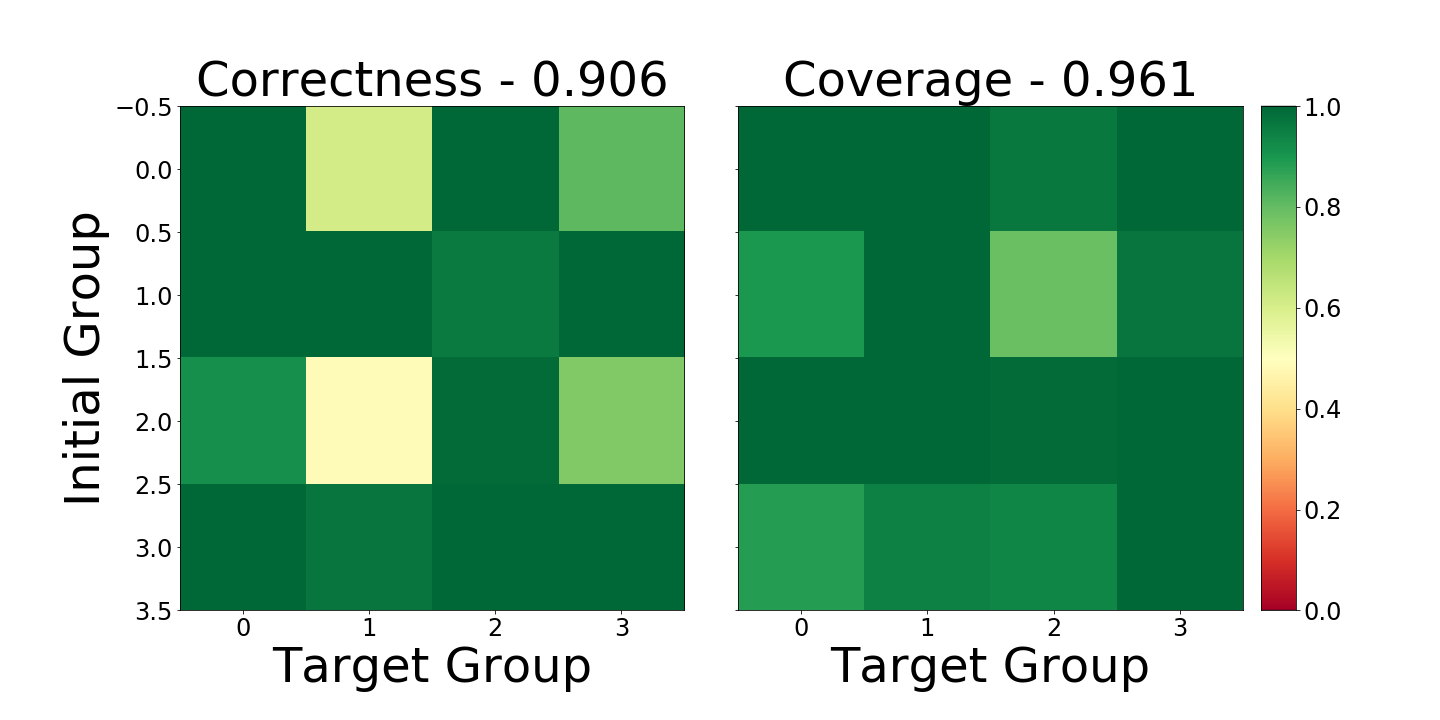}
	\caption{The pairwise metrics for \name on the synthetic dataset with no sparsity constraint.}
	\label{fig:a-synth-metrics}
	\includegraphics[width = 0.5\linewidth]{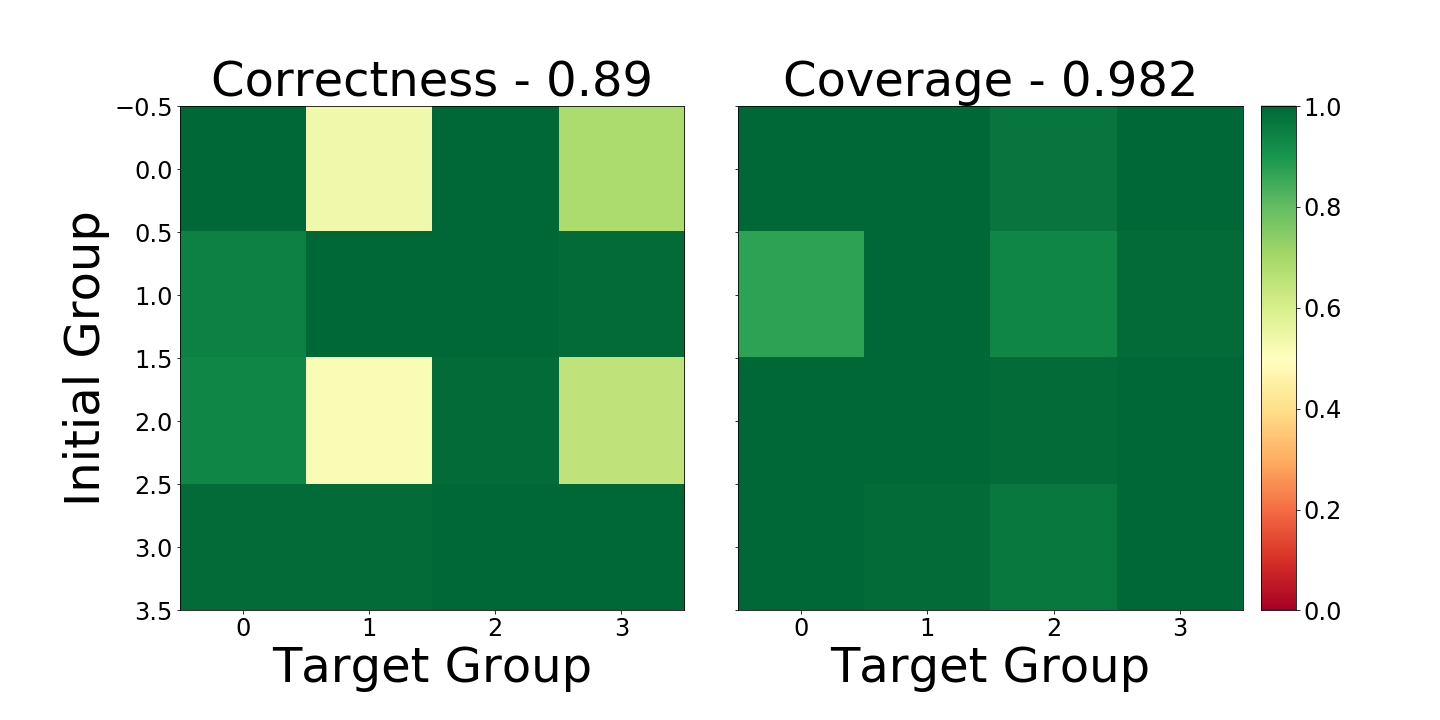}
	\caption{The pairwise metrics for \base on the synthetic dataset with no sparsity constraint.}
	\label{fig:a-synth-metrics-mean}
\end{figure}

\name is much better than \base for finding 250-sparse explanations on the single-cell RNA dataset (Figures \ref{fig:a-bipolar-metrics} and \ref{fig:a-bipolar-metrics-mean}).  
On the synthetic dataset, \name and \base are equally effective explanations of the model (Figures \ref{fig:a-synth-metrics} and \ref{fig:a-synth-metrics-mean}).  
However, \name only relied on the two causal variables while \base included the correlated variable as well.  

\subsection{Qualitative Analysis of the UCI Datasets using the Labels}
\label{appendix:uci}

Although the representations we learned for these datasets were trained in an unsupervised manner, the groups that they find often have strong connections to the labels for the datasets;  see Table \ref{fig:a-iris-labels}, Figure \ref{fig:a-housing-labels}, and Table \ref{fig:a-heart-labels}.  
By using the connection between the groups and labels, we will be able to qualitatively assess whether or not \name is finding real patterns in the data.  


\textbf{Iris Dataset.}
Looking at Table \ref{fig:a-iris-labels}, we can see that the groups in this representation match very closely with the class labels.
As a result, we would like to know whether or not the explanations \name finds are consistent with a model trained directly to predict the labels.  
For this comparison, we used a simple decision tree, which is shown in Figure \ref{fig:a-iris-tree}.
Looking at \name's explanations (Figure \ref{fig:a-iris-exp}), we can see that they largely agree with the decision tree since both primarily use Petal Width to separate the classes/groups.  

\begin{minipage}{0.45\linewidth}
	\centering
	\captionof{table}{The distribution of the labels per group for the UCI Iris dataset (classification).}
	\label{fig:a-iris-labels}
	\resizebox{\linewidth}{!}{
		\begin{tabular}{@{}l|lll@{}}
			\toprule
			\diagbox[width=10em]{Group}{Class} & Iris Setosa & Iris Versicolour & Iris Virginica \\ \midrule
			0 & 0 & 5 & 38 \\
			1 & 0 & 44 & 12 \\
			2 & 48 & 0 & 0 \\ \bottomrule
	\end{tabular}}
\end{minipage}%
\hspace{8pt}
\begin{minipage}{0.45 \linewidth}
	\centering
	\includegraphics[width = \linewidth]{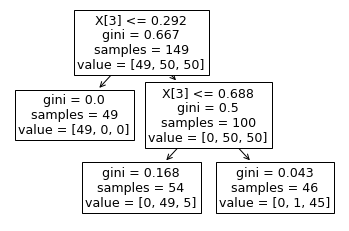}  
	\captionof{figure}{A small decision tree trained on this dataset.
		Notice that it relies on the Petal Width feature.}
	\label{fig:a-iris-tree} 
\end{minipage}

\begin{minipage}{.45\linewidth}
    \centering
    \includegraphics[width = \linewidth]{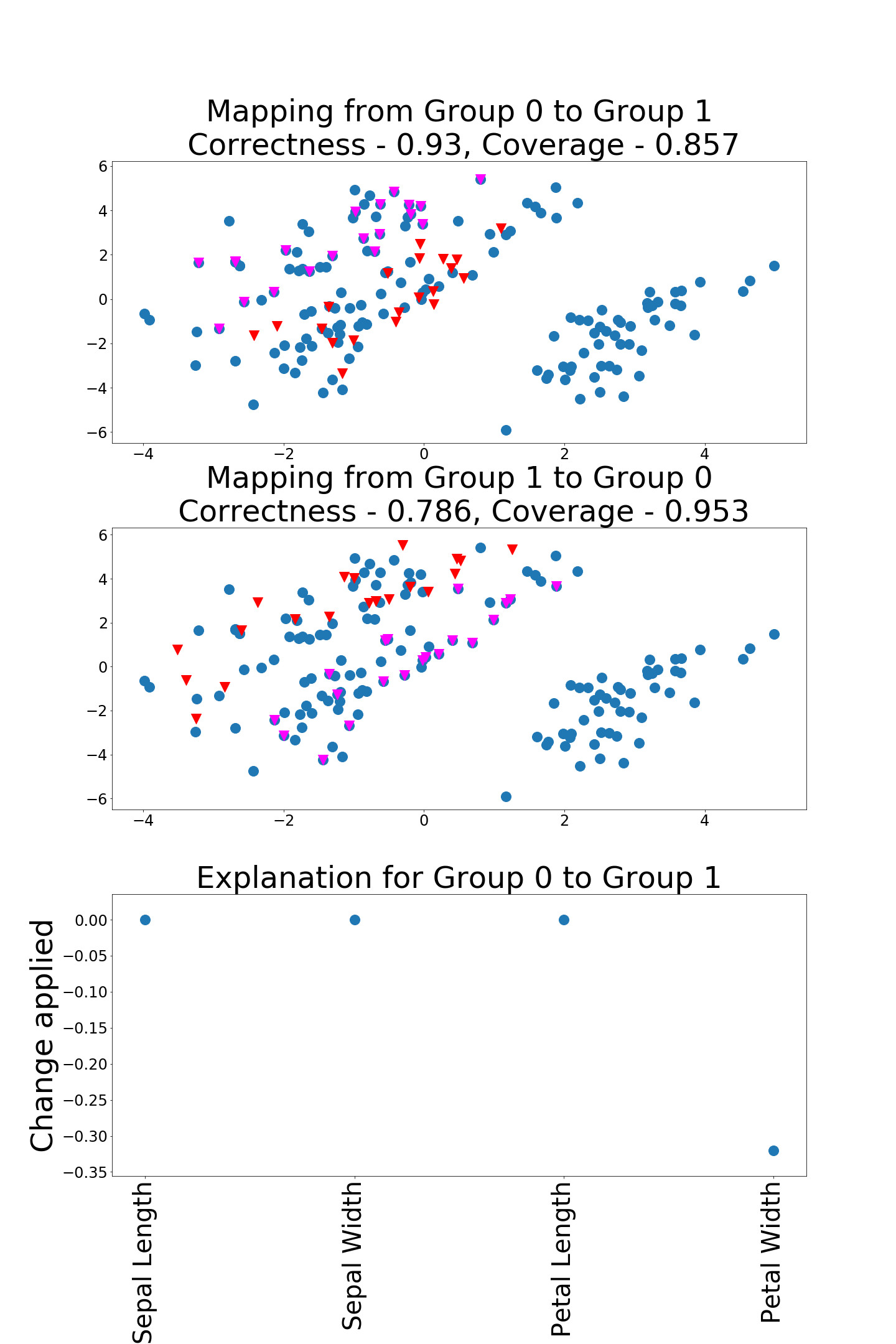}
\end{minipage}%
\hspace{8pt}
\begin{minipage}{.45\linewidth}
	\centering
	\includegraphics[width = \linewidth]{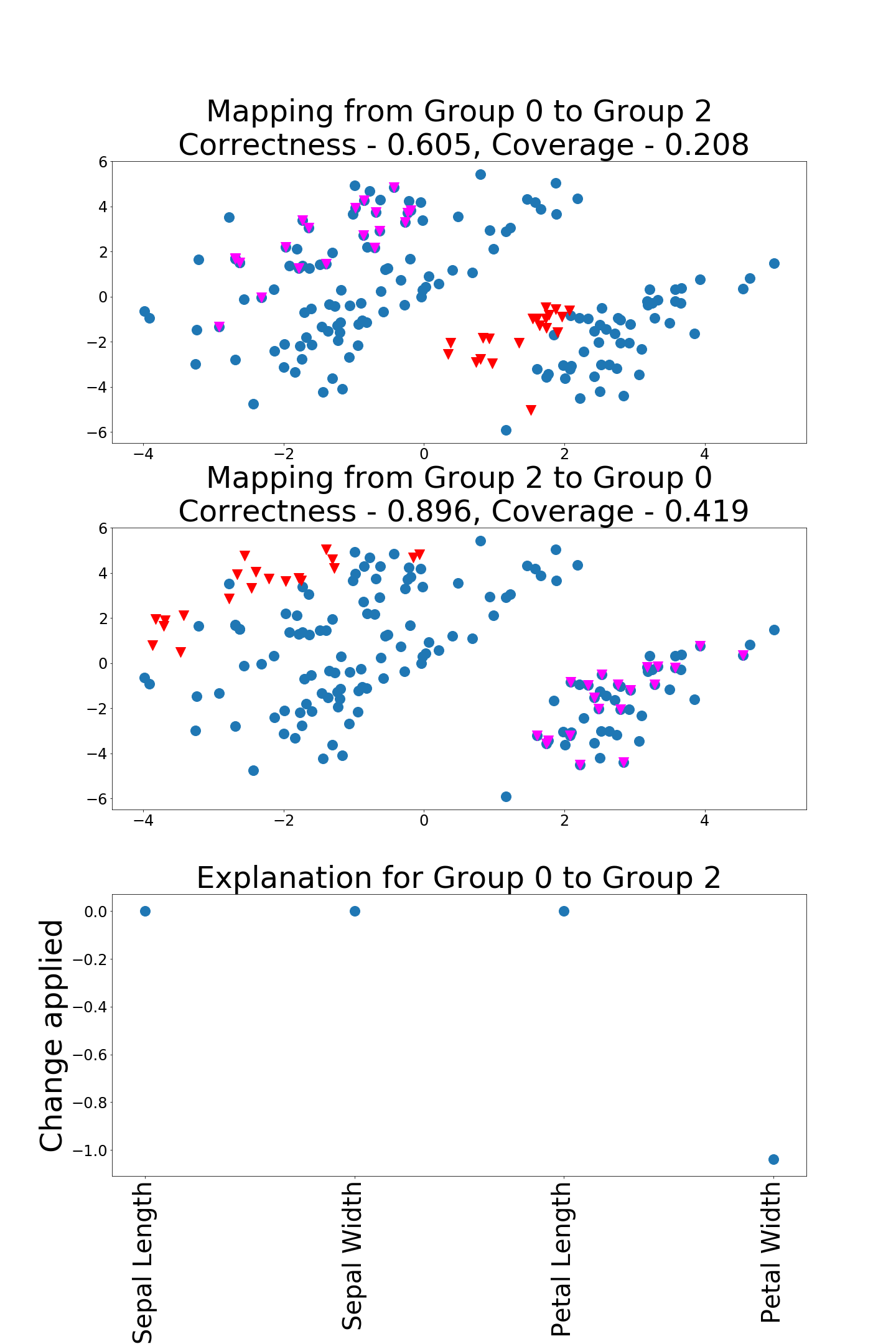}
\end{minipage}
\captionof{figure}{\name's 1-sparse explanation of the difference between Group 0 and Group 1 (left) and Group 0 and Group 2 (right).
Similar to the decision tree, they rely on the Petal Width feature.}
\label{fig:a-iris-exp}

\vspace{10pt}
\textbf{Boston Housing Dataset.}
Looking at Figure \ref{fig:a-housing-labels}, we can see that two comparisons between the groups stand out:  Group 0 to Group 2, which shows a significant increase in the price, and Group 3 to Group 5, which has relatively little effect on the price.  
As a result, we would like to determine what the differences between these groups of houses are that influence their price.

\begin{minipage}{\linewidth}
	\centering
	\includegraphics[width = 0.5\linewidth]{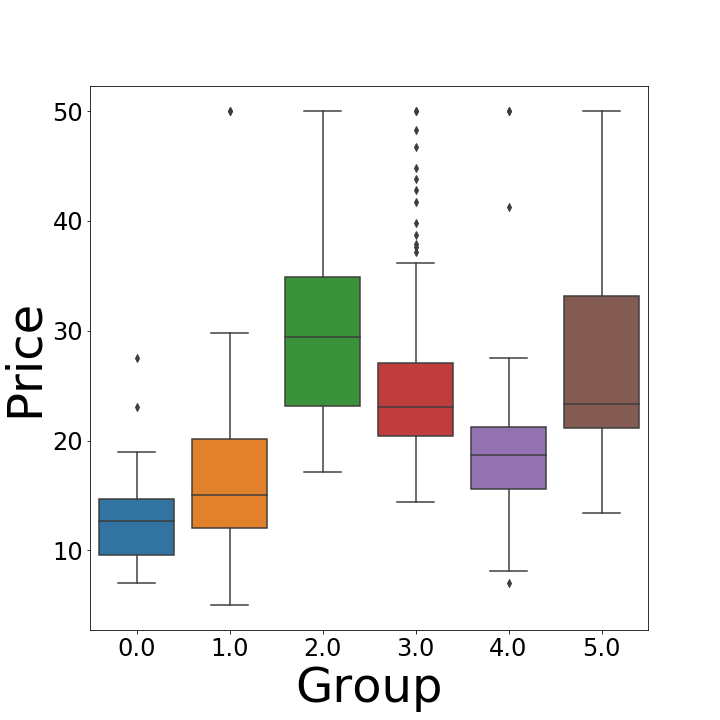}
	\captionof{figure}{The distribution of the labels per group for the UCI Boston Housing dataset (regression).}
	\label{fig:a-housing-labels}
\end{minipage} 
 
 Looking at Figure \ref{fig:a-housing-exp}, we see that \name found that the key differences between Group 0 and Group 2 appears to be the difference between a house being in an urban area vs being in a suburb:  the proportion of land zoned for large residential lots and B\footnote{%
This is a  unusually defined feature that is related to the racial demographics of a town.  
Determining what it means to change this feature depends on a measurement that is not in the dataset.} 
both increase while access to radial highways and tax rates both decrease.
It also found that the key differences between Group 3 and Group 5 are, first, moving the house onto the Charles river and, second, decreasing B. 

\begin{minipage}{.45\linewidth}
	\centering
	\includegraphics[width = \linewidth]{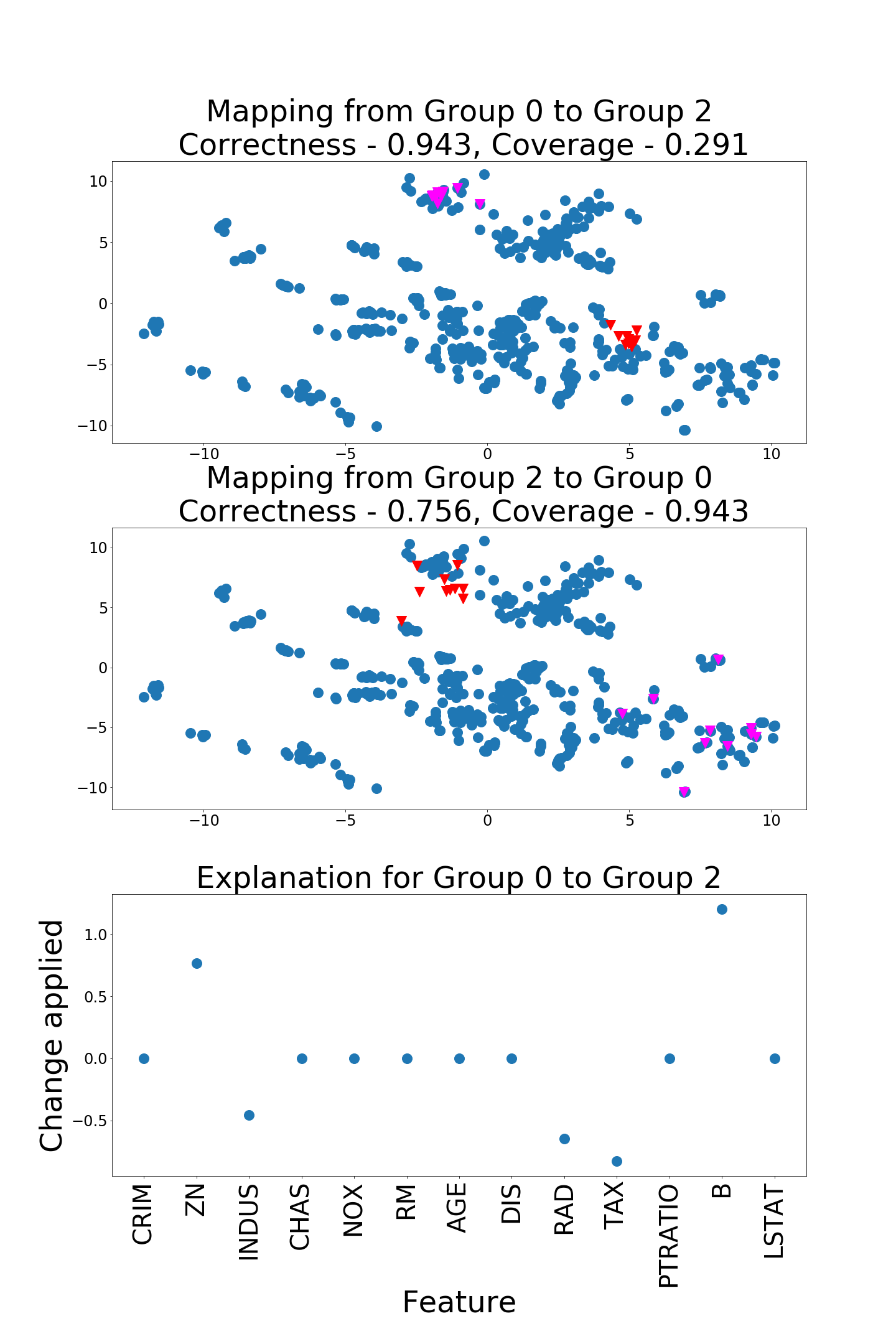}
\end{minipage}%
\hspace{8pt}
\begin{minipage}{.45\linewidth}
	\centering
	\includegraphics[width = \linewidth]{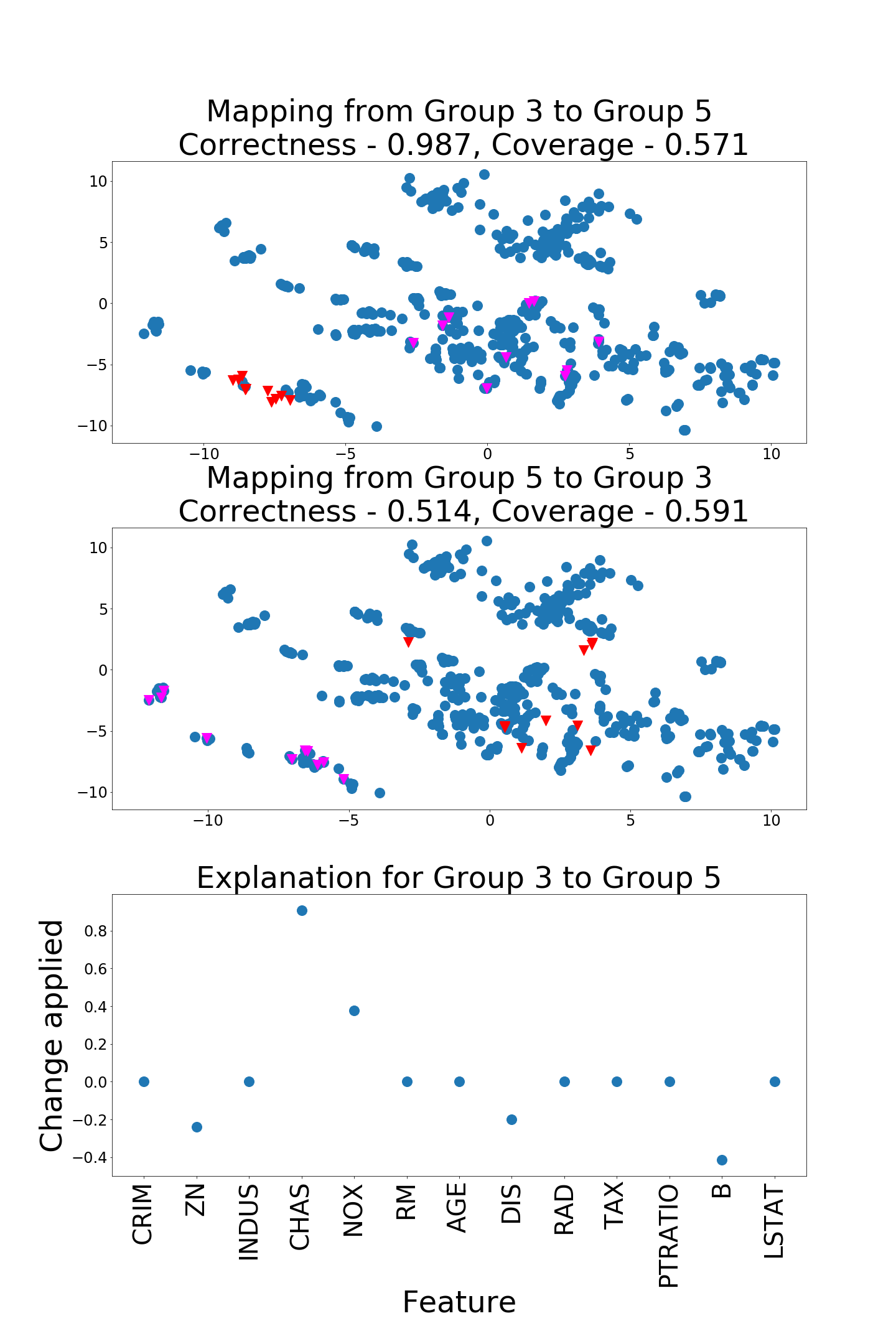}
\end{minipage}
\captionof{figure}{\name's 5-sparse explanation for Group 0 to Group 2 (Left) and Group 3 to Group 5 (right).}
\label{fig:a-housing-exp}

\vspace{10pt}
\textbf{Heart Disease Dataset.}
Looking at Figure \ref{fig:a-heart-labels}, we see that there are three large groups that stand out:  Group 1, which has a balanced risk of heart disease, Group 3, which has a relatively low risk, and Group 6, which has a relatively high risk.  
As a result, we would like to determine what the differences between these groups of subjects are that influence their risk of heart disease.  

\begin{minipage}{\linewidth}
	\centering
	\captionof{table}{The distribution of the labels per group for the UCI Heart Disease dataset (classification).}
	\label{fig:a-heart-labels}
	\resizebox{0.5\linewidth}{!}{
		\begin{tabular}{@{}l|ll@{}}
			\toprule
			\diagbox[width=10em]{Group}{Class} & No Heart Disease & Heart Disease \\ \midrule
			0 & 6 & 15 \\
			1 & 46 & 62 \\
			2 & 10 & 1 \\
			3 & 52 & 14 \\
			4 & 4 & 1 \\
			5 & 10 & 7 \\
			6 & 8 & 59 \\
			7 & 2 & 5 \\ \bottomrule
	\end{tabular}}
\end{minipage}

Looking at Figure \ref{fig:a-heart-exp}, \name found that the key differences between Group 1 and Group 3 are a moderate decrease in chest pain along with having exercised induced angina;  these are subjects whose symptoms are explained by exercise induced angina rather than heart disease.  
It also found that the key difference between Group 1 and Group 6 is that Group 1 is made up of men while Group 6 is made up of women; this is consistent with the fact that heart disease is the leading cause of mortality in women \cite{bello2004epidemiology}.  

\begin{minipage}{.45\linewidth}
	\centering
	\includegraphics[width = \linewidth]{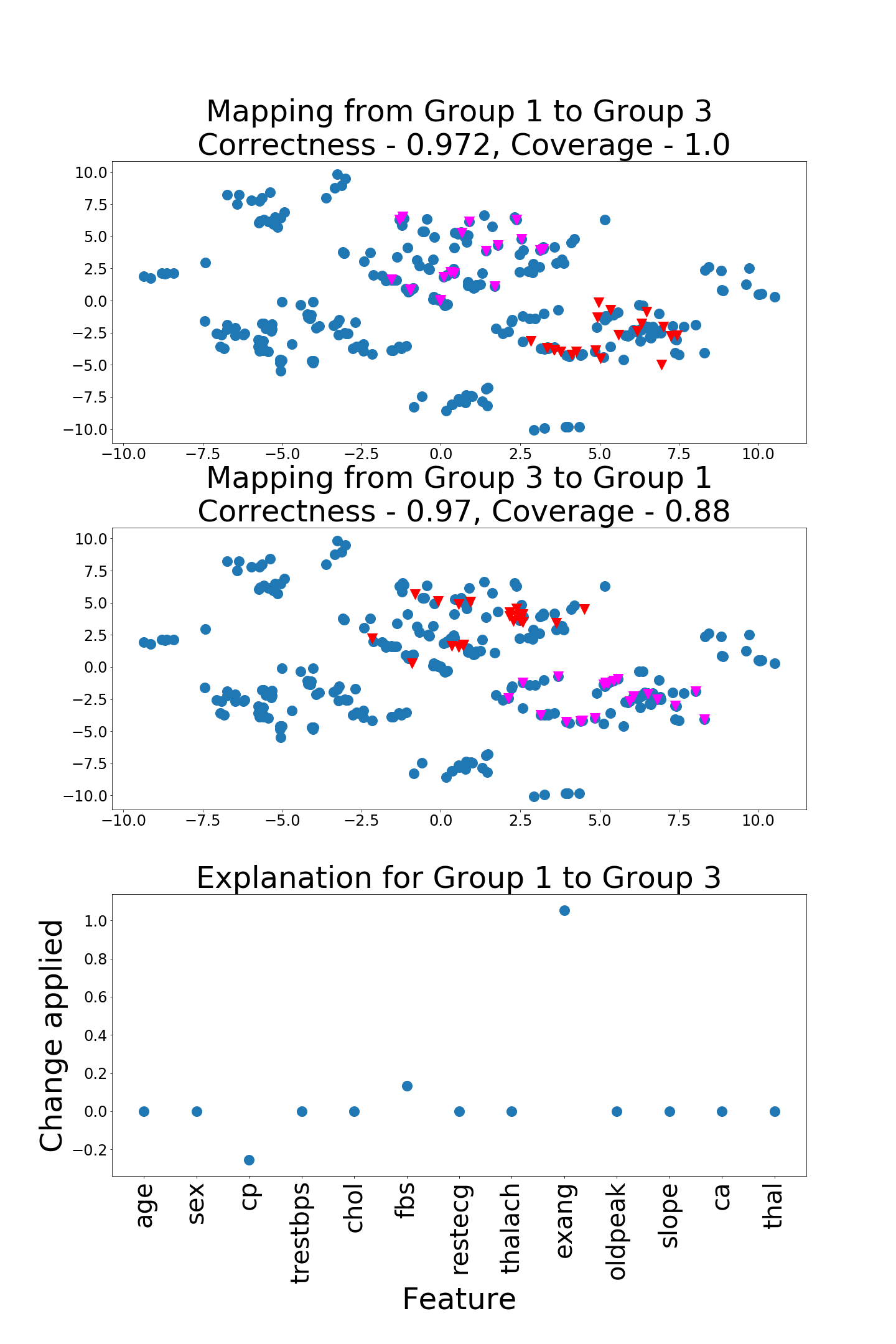}
\end{minipage}%
\hspace{8pt}
\begin{minipage}{.45\linewidth}
	\centering
	\includegraphics[width = \linewidth]{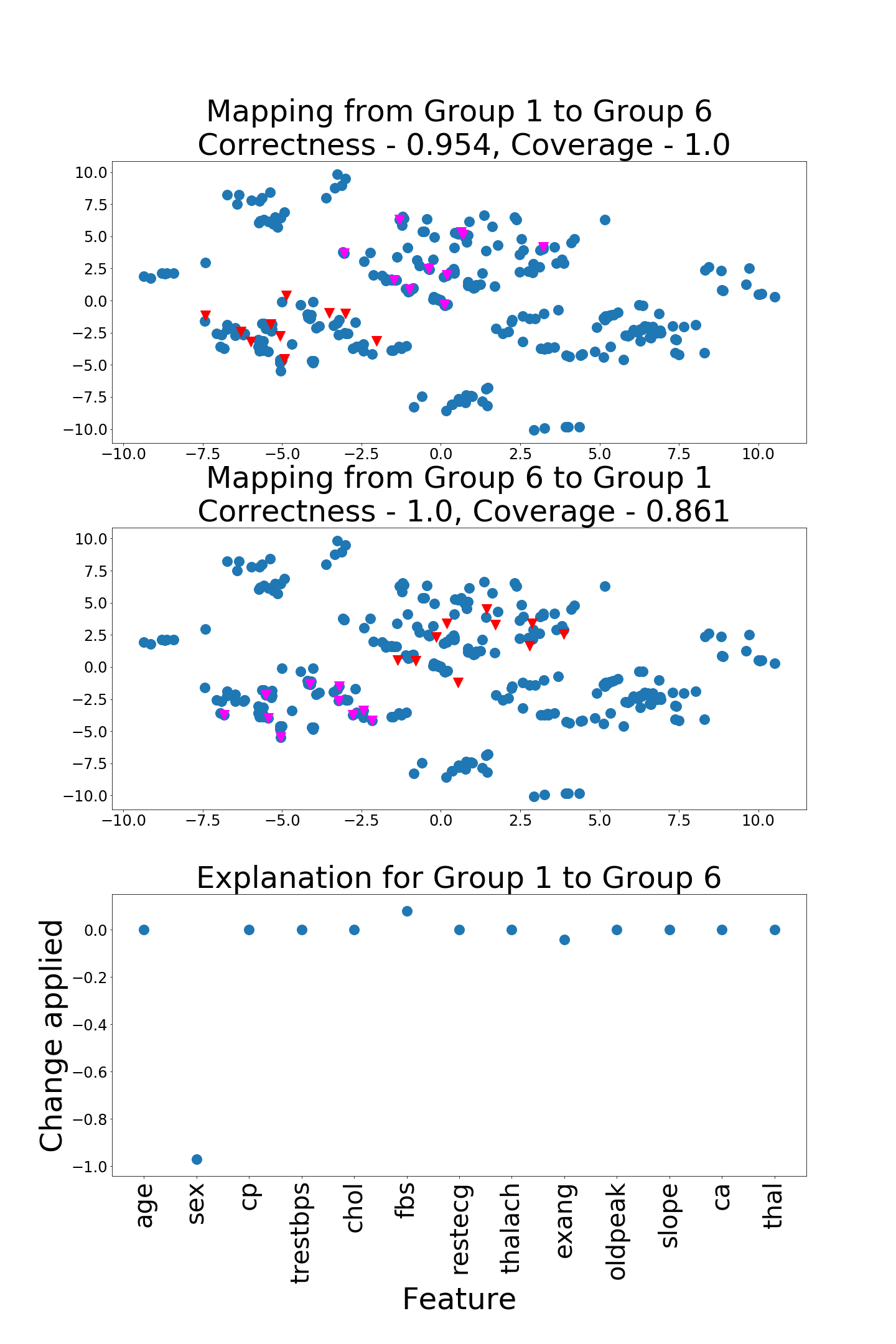}
\end{minipage}
\captionof{figure}{\name's 3-sparse explanations for the difference between Group 1 and Group 3 (Left) and Group 1 and Group 6 (right).}
\label{fig:a-heart-exp}

\subsection{Quantitative Analysis of Modified Versions of the UCI Datasets.}
\label{appendix:modify}

Because the UCI datasets are not synthetic datasets, we do not know the underlying process that generated the data and, as a result, it is difficult to quantitatively determine whether or not an explanation is ``correct'' in the way that we could with the synthetic dataset.
Consequently, we performed a series of experiments on modified versions of the original datasets in order to answer two important questions:  
\begin{itemize}
    \item Does \name correctly identify the modifications we made to the original dataset?
    \item Do \name's explanations between the original groups change when the modified group is added to the dataset? 
\end{itemize}
We found that \name does identify the modifications we made and that, in doing so, it does not significantly change the explanations between the original groups. 
Importantly, this result remains true even if we retrain the learned representation on the modified dataset.  
These results are a strong indicator that \name is finding real patterns in the data.  

\vspace{10pt}
\textbf{How do we modify the datasets?}
We create a modified version of the original dataset by:  picking one of the groups of points in the original dataset, modifying that group of points in some way, and adding that new modified group of points to the original dataset.
We will call the original dataset $D$ and the modified dataset $D'$, where $D'= D \cup G'$ and $G'$ is the modified version of some group of points $G$.  

The critical choice to make during this process is to determine what modification to apply to $G$ to get $G'$.  
We chose to add random noise to some of the features of the points in $G$ and used the following two criteria when defining this modification for a particular dataset:
\begin{itemize}
    \item $G'$ should be approximately within the range/distribution of $D$.
    \item $r(G')$ should form it's own (approximately) distinct group.
    Intuitively, if $r(G')$ does not form its own group, then $r$ thinks $G'$ is similar to some other group in the dataset and, as a result, we would not expect \name to be able to explain the differences between $G'$ and that group of points.
\end{itemize}
The modifications we used are in Table \ref{tab:a-pert}.  

\begin{table}[h]
\caption{For each dataset, we chose a group of points to modify and modified it by applying these perturbations to the specified features.}
\label{tab:a-pert}
\centering
\resizebox{\linewidth}{!}{
\begin{tabular}{@{}llll@{}}
\toprule
Dataset & Group Modified & Feature  & Perturbation Applied \\ \midrule
Iris & 0 & Sepal Width & -0.4 + Uniform(-0.1, 0.1) \\
Housing & 1 & ZN & 0.9 + Uniform(-0.1, 0.1) \\
 &  & TAX & -0.5 + Uniform(-0.1, 0.1) \\
Heart & 1 & restecg & -0.9 + Uniform(-0.1, 0.1) \\
 &  & exang & 0.6 + Uniform(-0.1, 0.1) \\ \bottomrule
\end{tabular}}
\end{table}

\vspace{8pt}
\textbf{Experimental Setup:}
We now have two versions of each dataset:  $D$ and $D'$.  
We also have the original learned representation $r$, which was trained on $D$, and a new learned representation $r'$, which is trained on $D'$.  
As a result, we have three sets of explanations:
\begin{itemize}
    \item \textbf{Original:} These explain $r$ when applied to $D$
    \item \textbf{Modified:} These explain $r$ when applied to $D'$
    \item \textbf{Retrained:} These explain $r'$ when applied to $D'$
\end{itemize}
The visualization of the representation for the first setting is in the Appendix \ref{appendix:rep} and the later two these settings is in Figures \ref{fig:a-iris}, \ref{fig:a-housing}, and \ref{fig:a-heart}.  
Note that applying $r$ to $D'$ looks the same as applying $r$ to $D$ except for the fact that there is an additional group from adding $G'$ to $D$ and that applying $r'$ to $D'$ often shows that $r'$ has learned to separate $G'$ from the other groups better than $r$ did.  

\begin{minipage}{.45\linewidth}
	\centering
	\includegraphics[width = \linewidth]{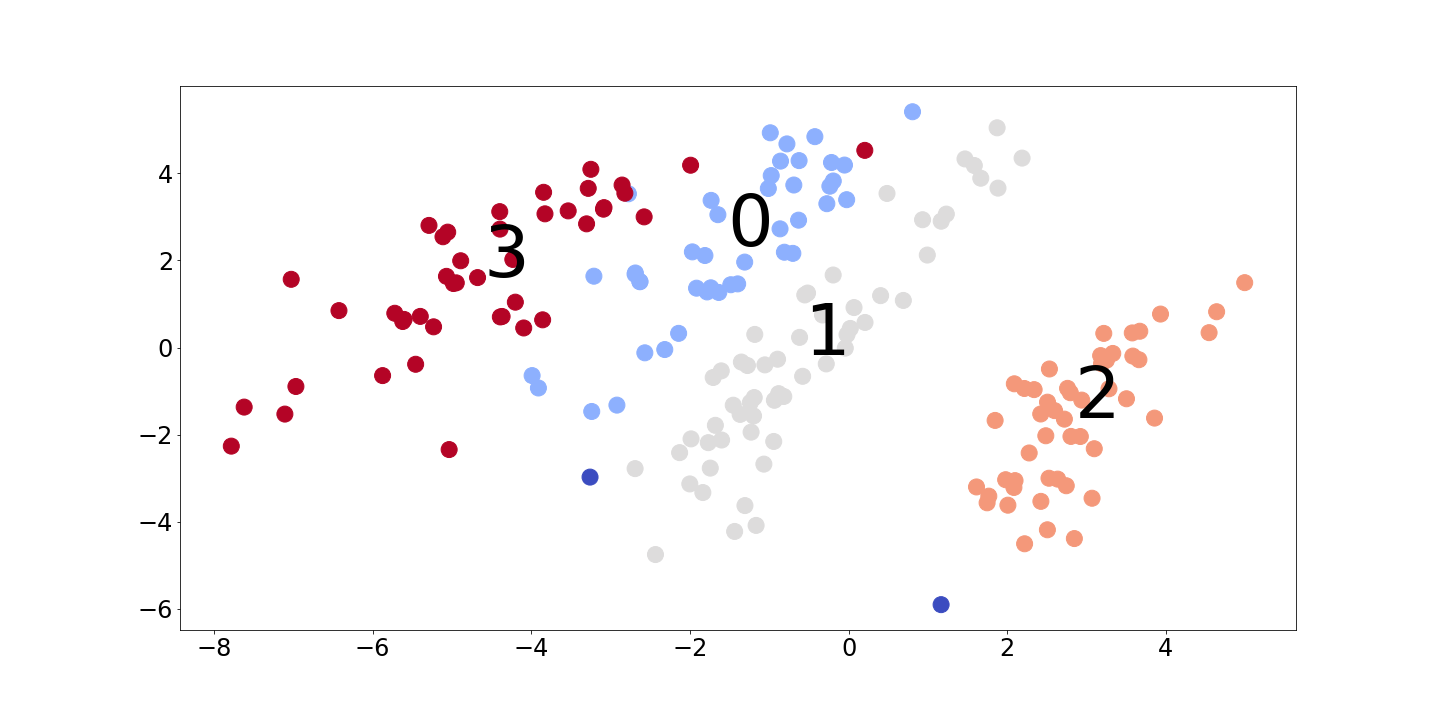} 
\end{minipage}%
\hspace{8pt}
\begin{minipage}{.45\linewidth}
	\centering
	\includegraphics[width = \linewidth]{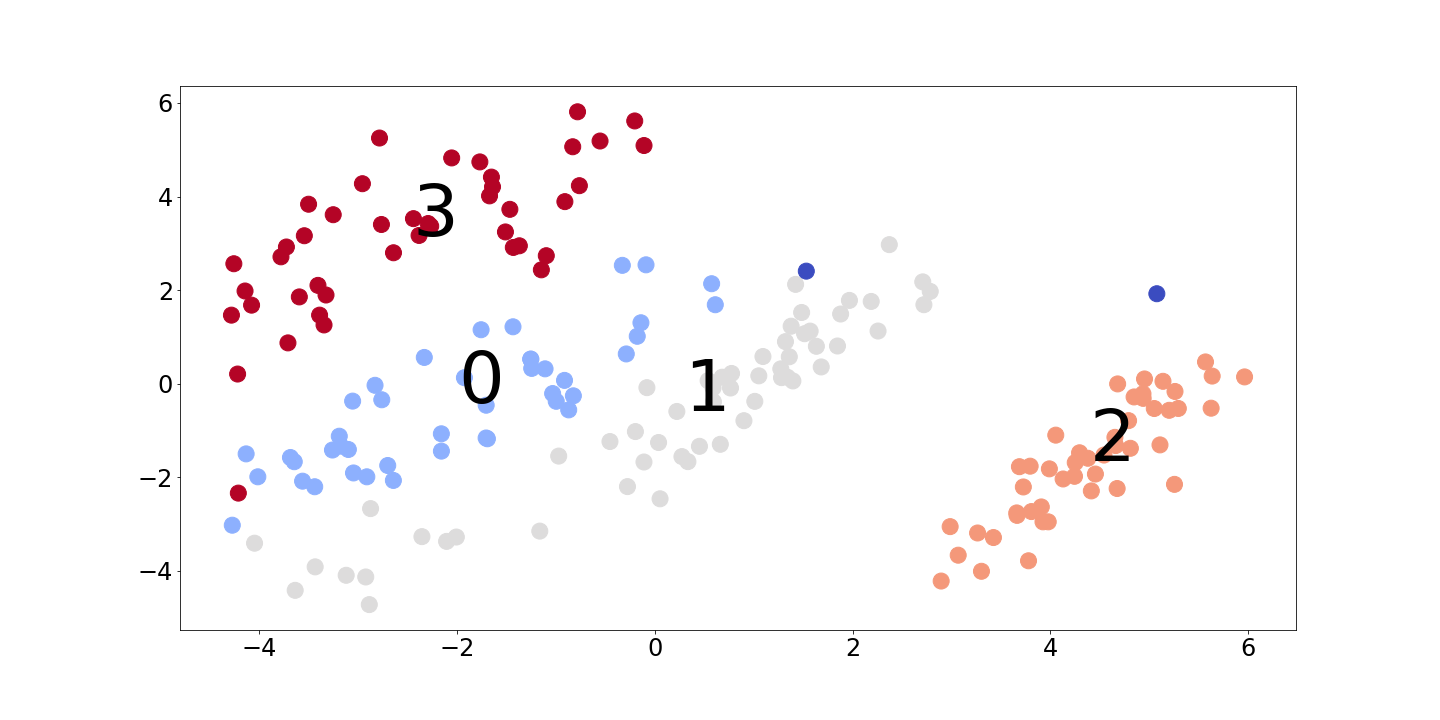}
\end{minipage}
\captionof{figure}{The learned representation for:  $r$ applied to $D'$ (Left) and $r'$ applied to $D'$ (Right) for the Iris dataset}
\label{fig:a-iris}

\begin{minipage}{.45\linewidth}
	\centering
	\includegraphics[width = \linewidth]{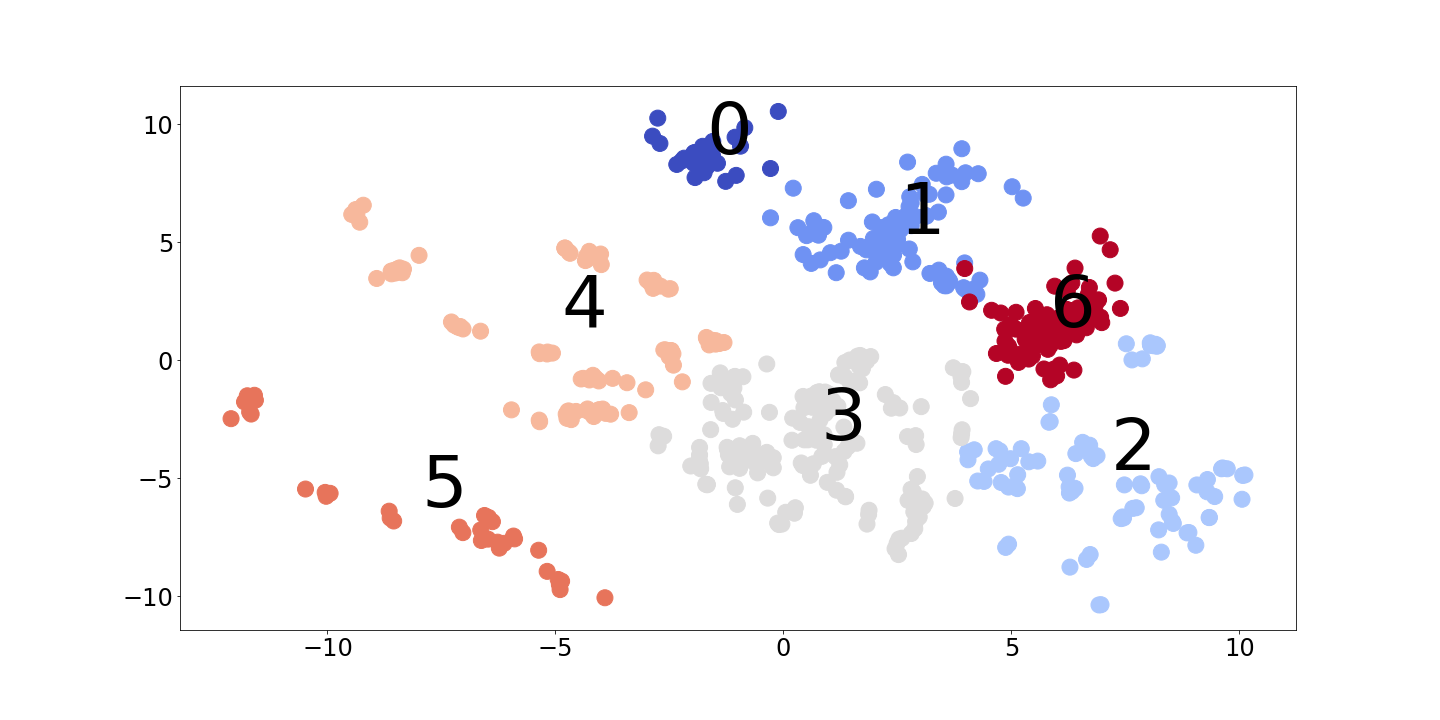} 
\end{minipage}%
\hspace{8pt}
\begin{minipage}{.45\linewidth}
	\centering
	\includegraphics[width = \linewidth]{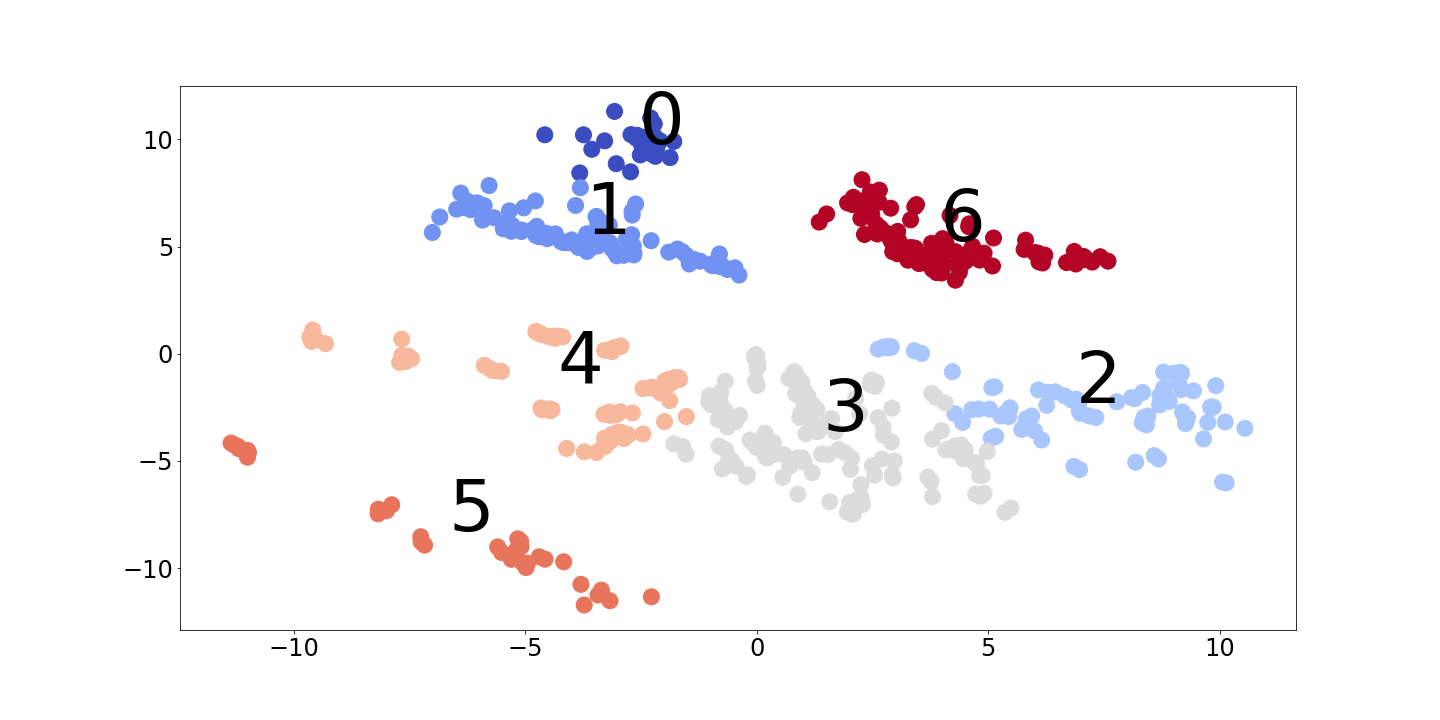}
\end{minipage}
\captionof{figure}{The learned representation for:  $r$ applied to $D'$ (Left) and $r'$ applied to $D'$ (Right) for the Boston Housing dataset}
\label{fig:a-housing}

\vspace{30pt}

\begin{minipage}{.45\linewidth}
	\centering
	\includegraphics[width = \linewidth]{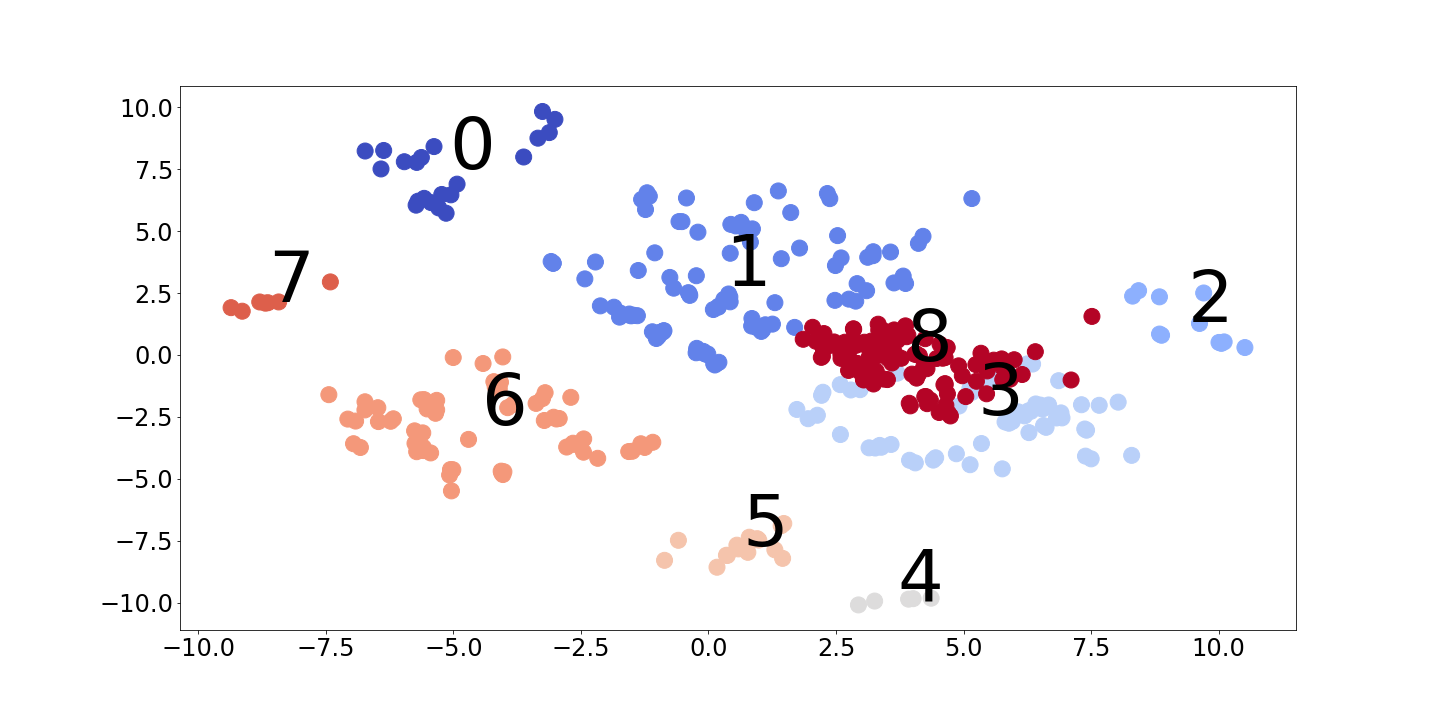} 
\end{minipage}%
\hspace{8pt}
\begin{minipage}{.45\linewidth}
	\centering
	\includegraphics[width = \linewidth]{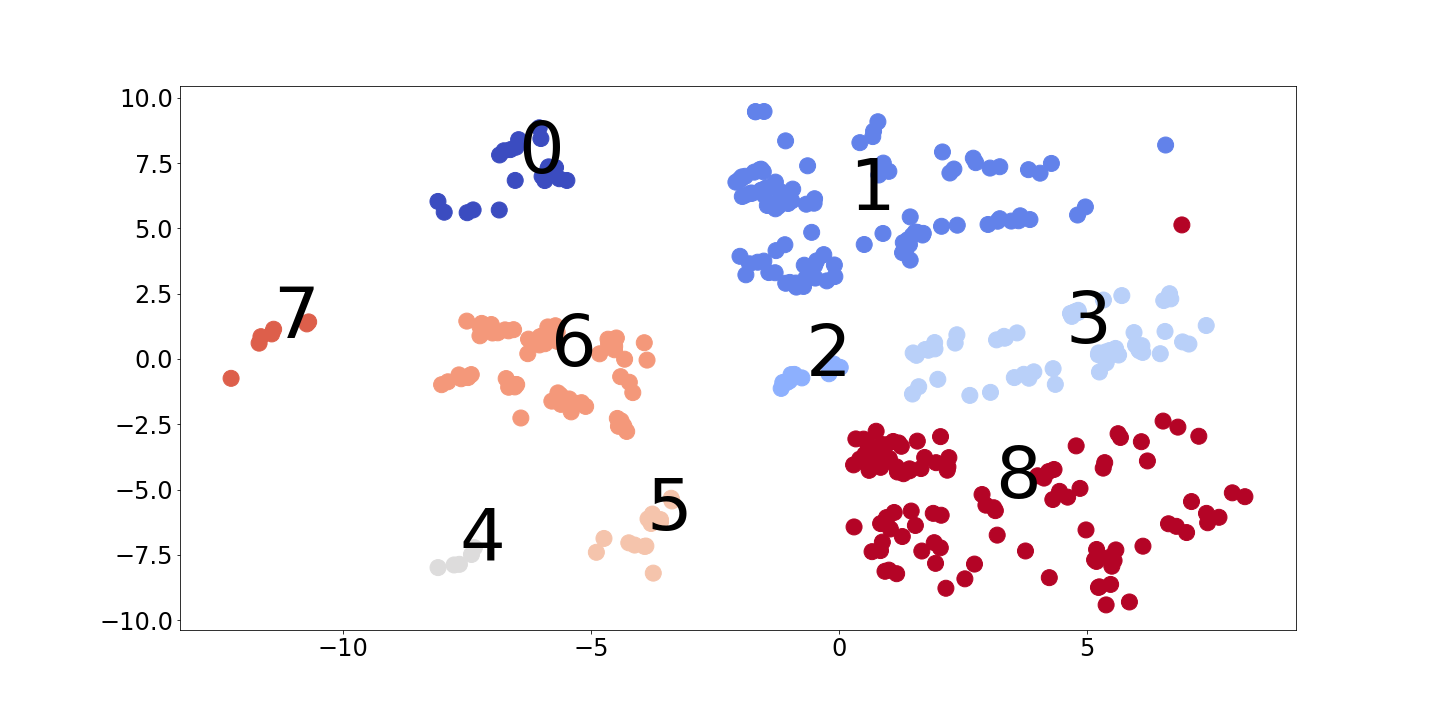}
\end{minipage}
\captionof{figure}{The learned representation for:  $r$ applied to $D'$ (Left) and $r'$ applied to $D'$ (Right) for the Heart Disease dataset}
\label{fig:a-heart}

\vspace{8pt}
\textbf{Does \name correctly identify the modifications we made to the original dataset?}
In Figure  Figures \ref{fig:a-iris-recover}, \ref{fig:a-housing-recover}, and \ref{fig:a-heart-recover}, we can see the explanations \name found for the difference between $G$ and $G'$ for each of the datasets.  
If we compare the explanations to the modifications from Table \ref{tab:a-pert}, we can see that they identified which features we changed and, approximately, by how much.  
The error in the estimation of ``by how much'' is due to the $l_1$ regularization used to find a simple explanation. 

\begin{minipage}{.45\linewidth}
	\centering
	\includegraphics[width = \linewidth]{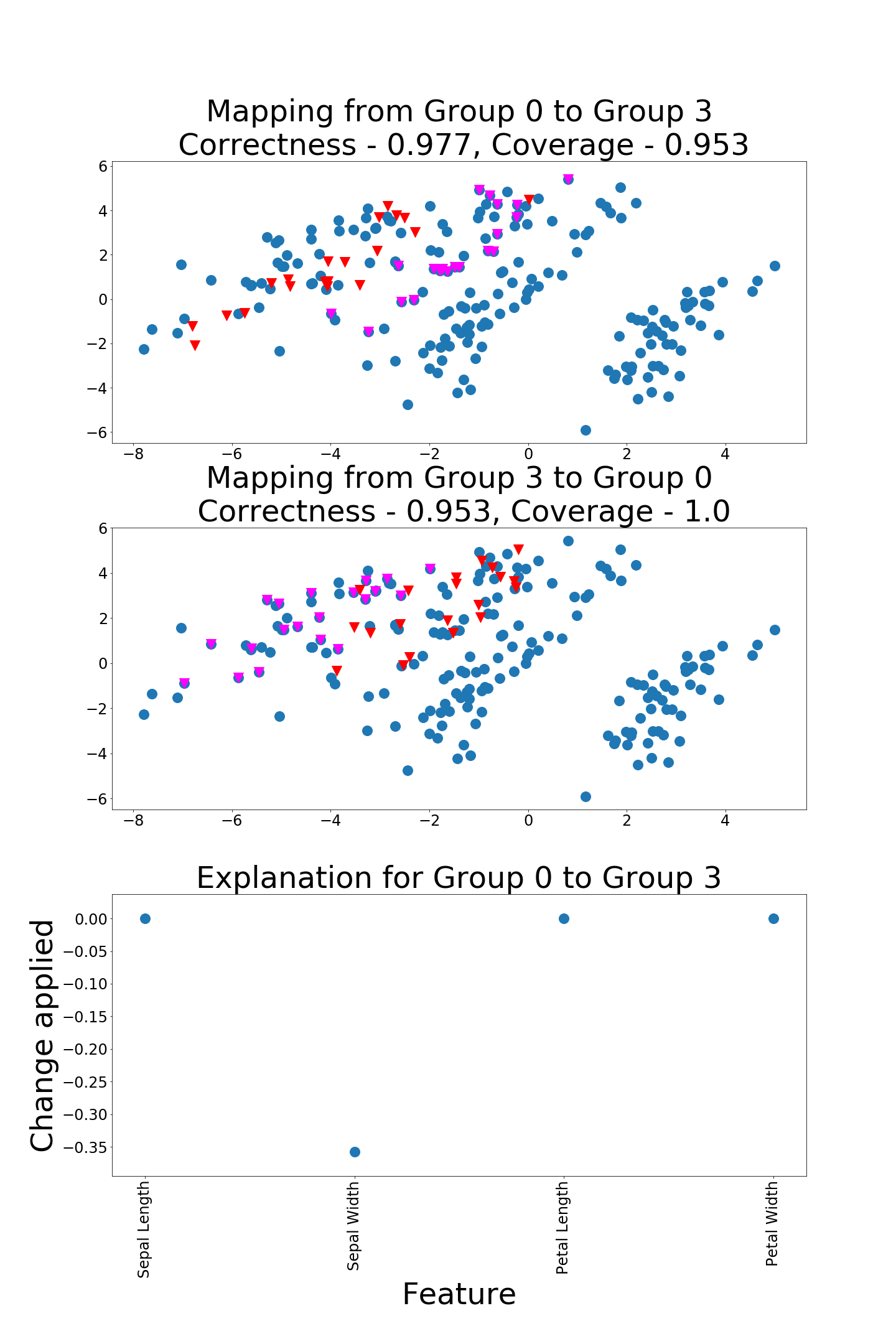} 
\end{minipage}%
\hspace{8pt}
\begin{minipage}{.45\linewidth}
	\centering
	\includegraphics[width = \linewidth]{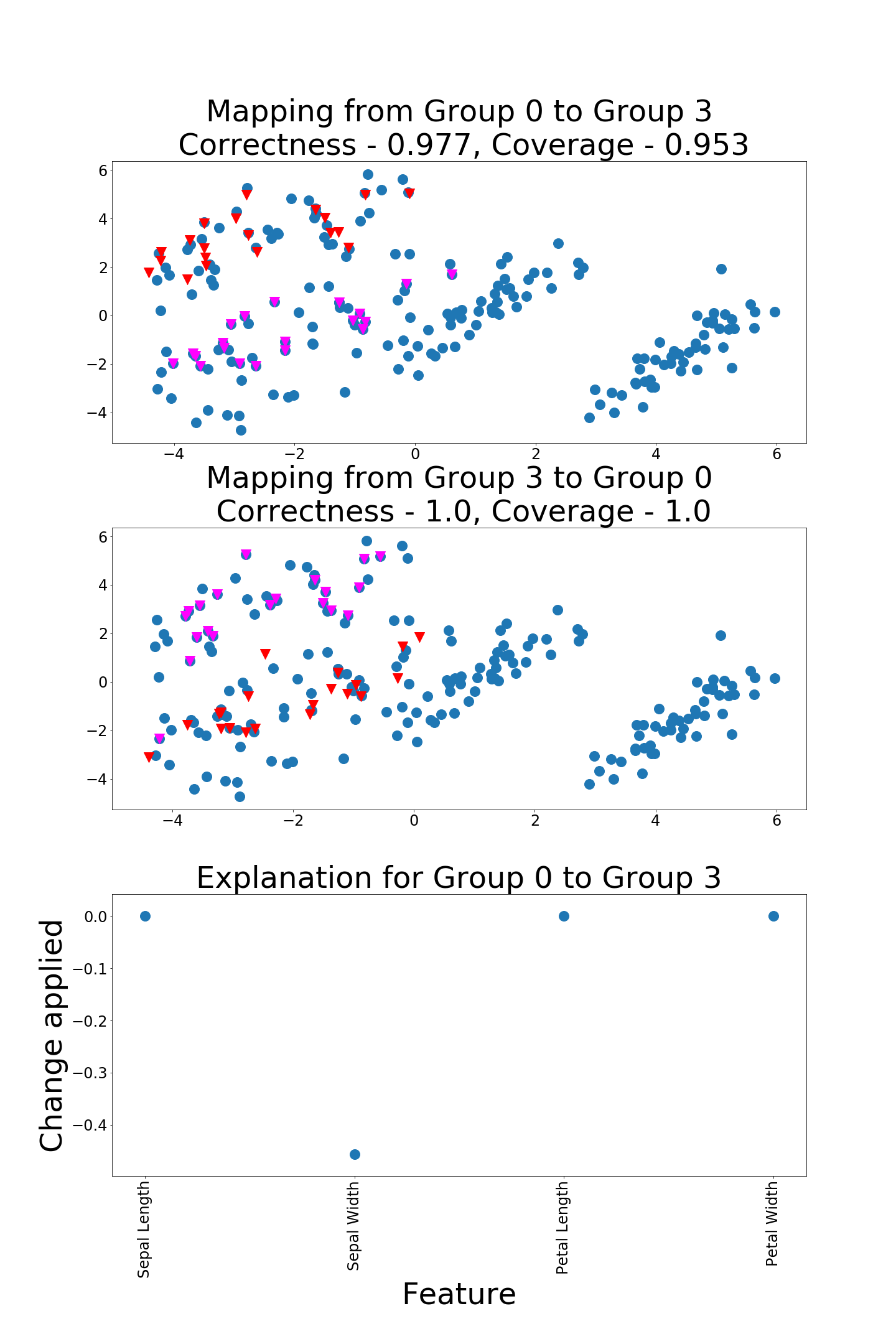}
\end{minipage}
\captionof{figure}{The Modified (Left) and Retrained (Right) explanation's explanation for the difference between $G$ and $G'$ on the Iris dataset.}
\label{fig:a-iris-recover}

\begin{minipage}{.45\linewidth}
	\centering
	\includegraphics[width = \linewidth]{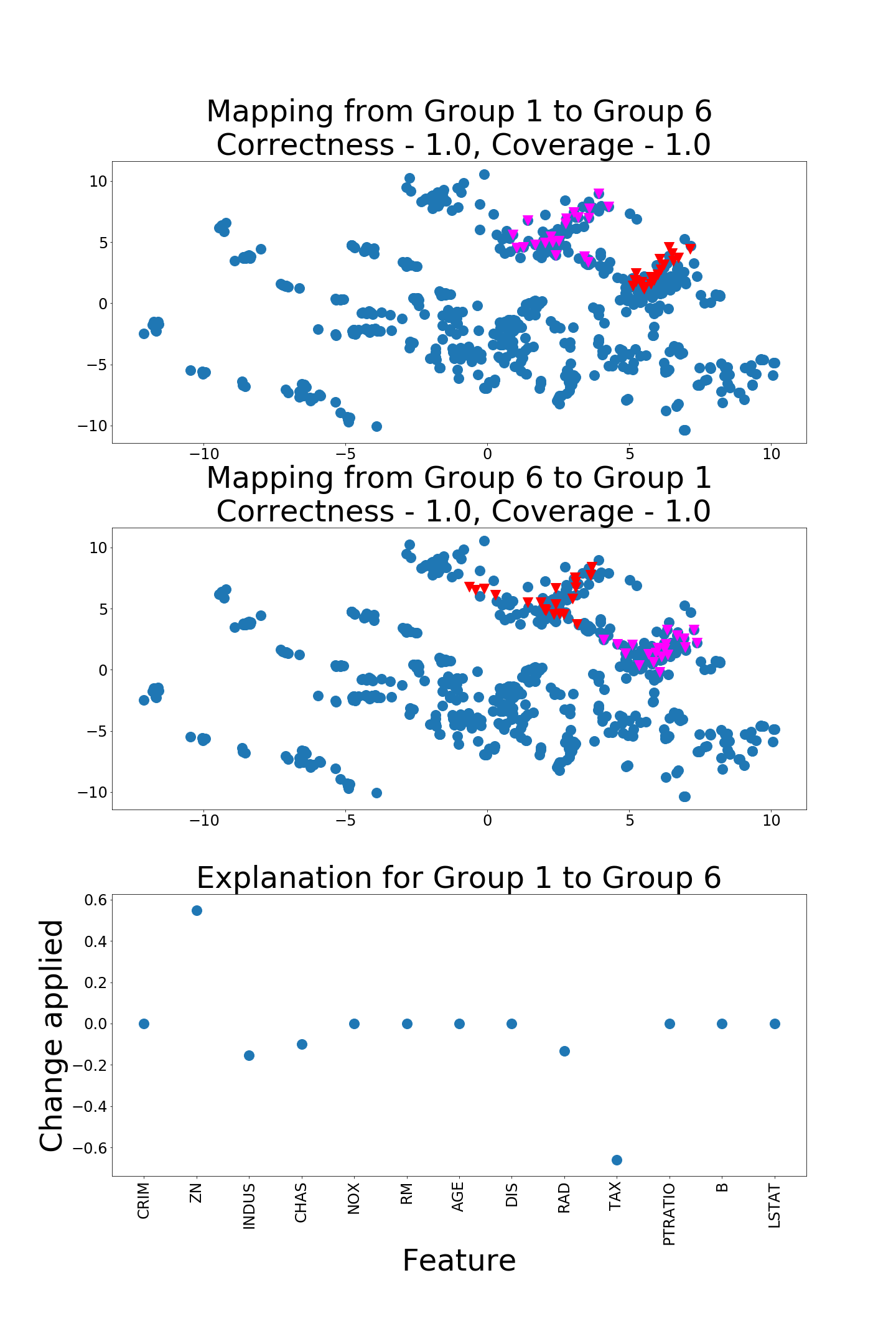} 
\end{minipage}%
\hspace{8pt}
\begin{minipage}{.45\linewidth}
	\centering
	\includegraphics[width = \linewidth]{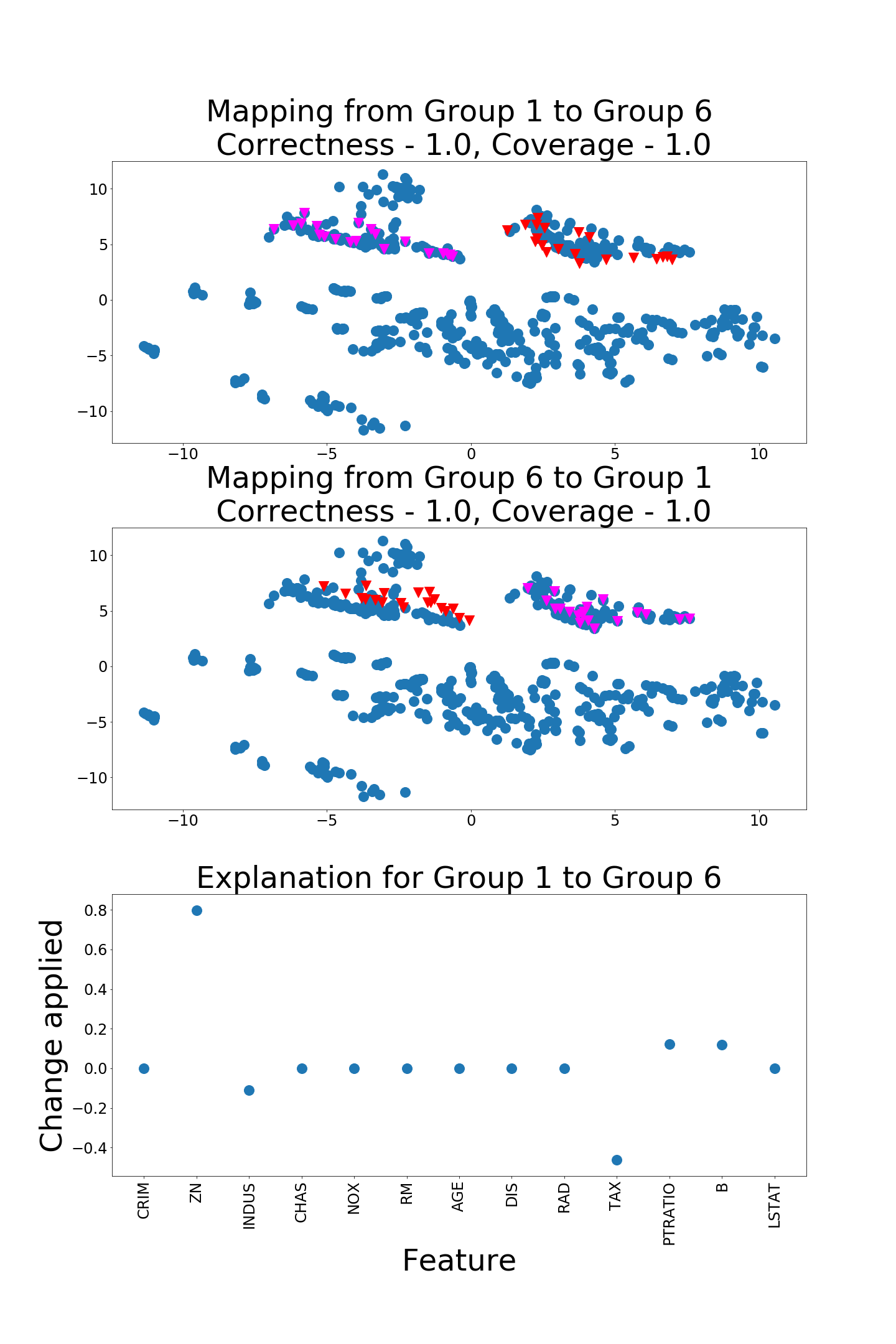}
\end{minipage}
\captionof{figure}{The Modified (Left) and Retrained (Right) explanation's explanation for the difference between $G$ and $G'$ on the Boston Housing dataset.}
\label{fig:a-housing-recover}

\begin{minipage}{.45\linewidth}
	\centering
	\includegraphics[width = \linewidth]{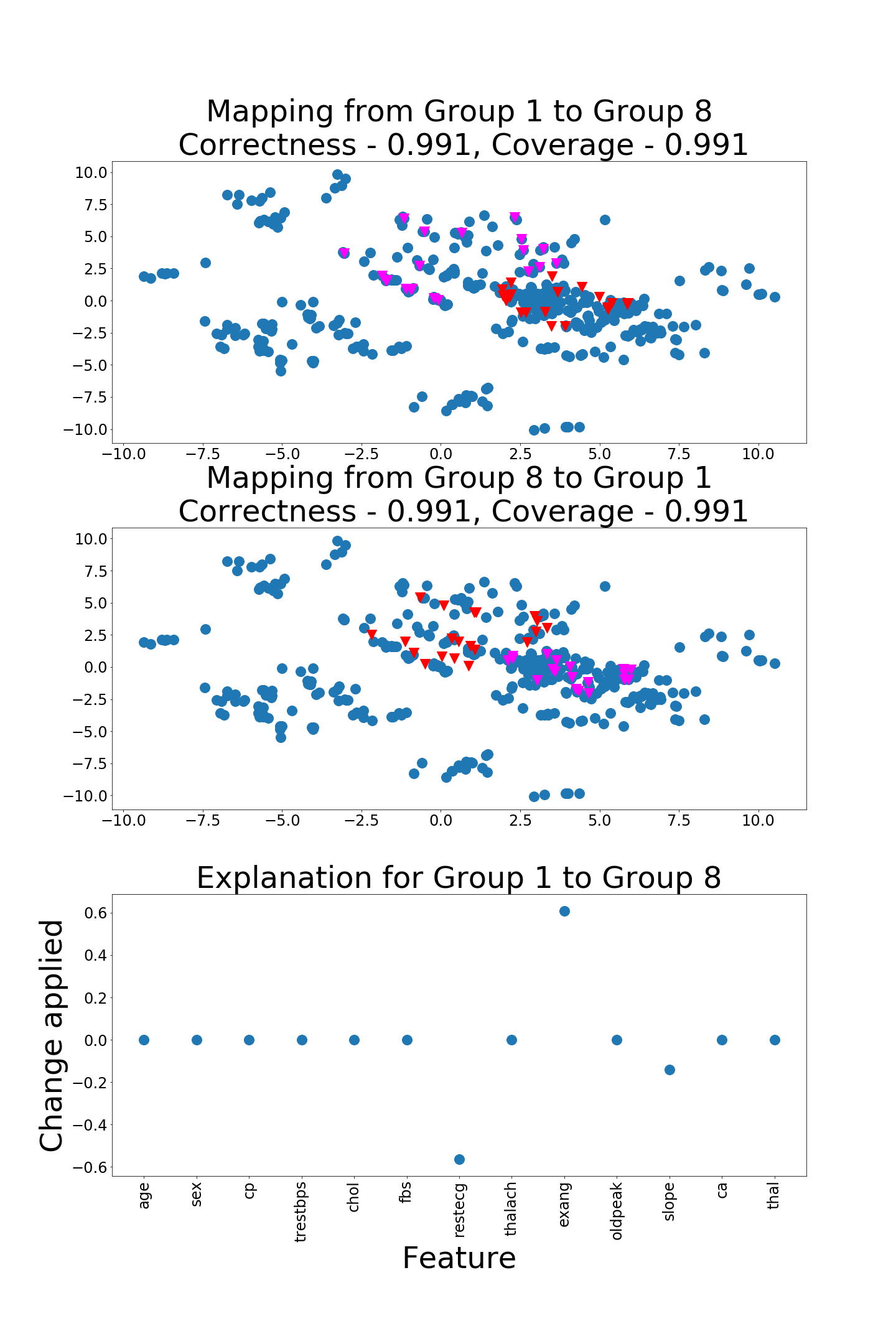} 
\end{minipage}%
\hspace{8pt}
\begin{minipage}{.45\linewidth}
	\centering
	\includegraphics[width = \linewidth]{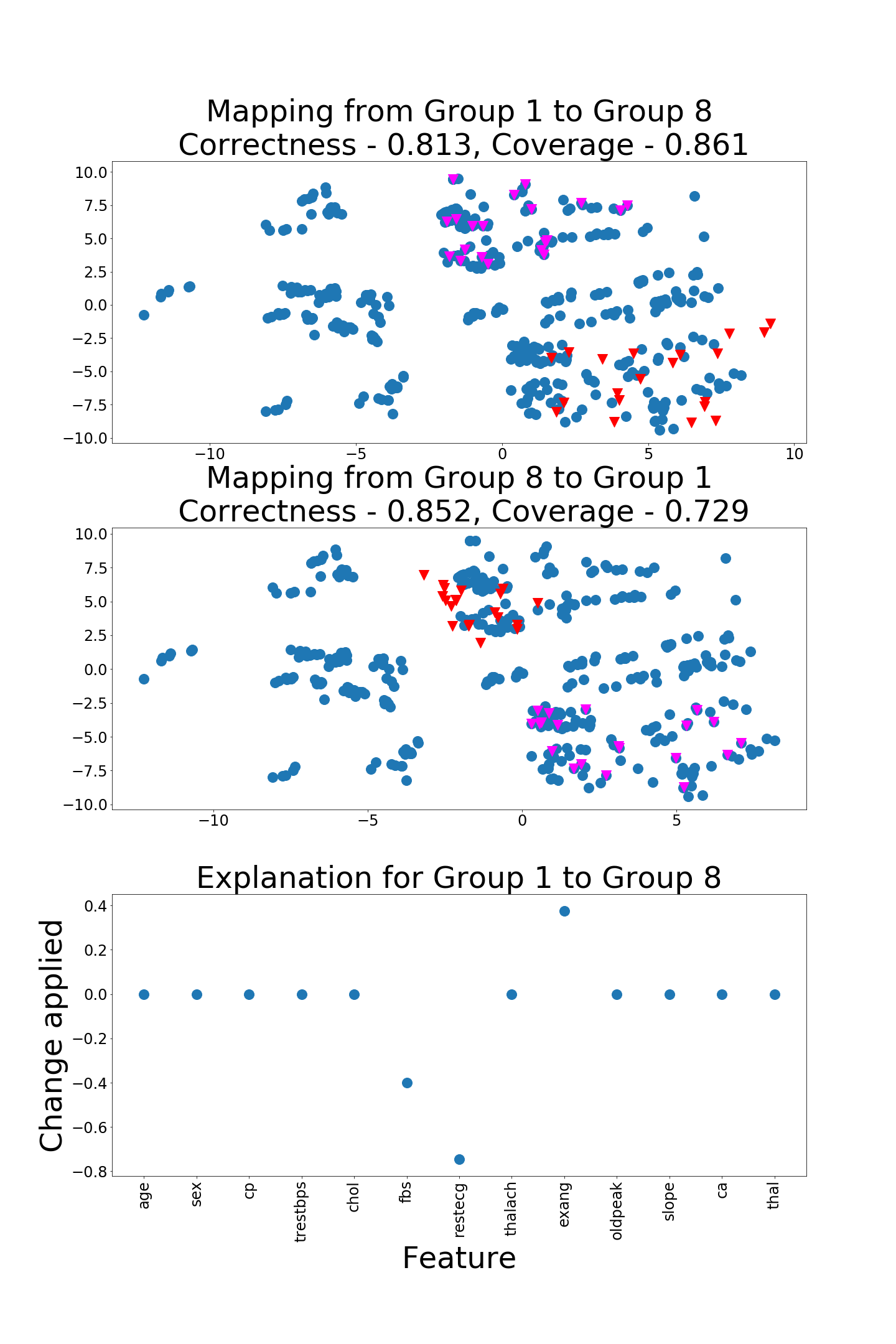}
\end{minipage}
\captionof{figure}{The Modified (Left) and Retrained (Right) explanation's explanation for the difference between $G$ and $G'$ on the Heart Disease dataset.}
\label{fig:a-heart-recover}

\vspace{8pt}
\textbf{Do \name's explanations between the original groups change when the modified group is added to the dataset?}
In Figures \ref{fig:a-iris-similarity}, \ref{fig:a-housing-similarity}, and \ref{fig:a-heart-similarity}, we can see a comparison of the explanations for the differences between the original groups for the Original vs the Modified and the Original vs the Retrained explanations.  
Adding $G'$ to $D$ did not cause \name to find significantly different explanations between the groups in $D$.  
Explaining $r'$ resulted in explanations that were generally similar, but adding another layer of variability (\ie training $r'$) did add some noise.  

\begin{minipage}{.45\linewidth}
	\centering
	\includegraphics[width = \linewidth]{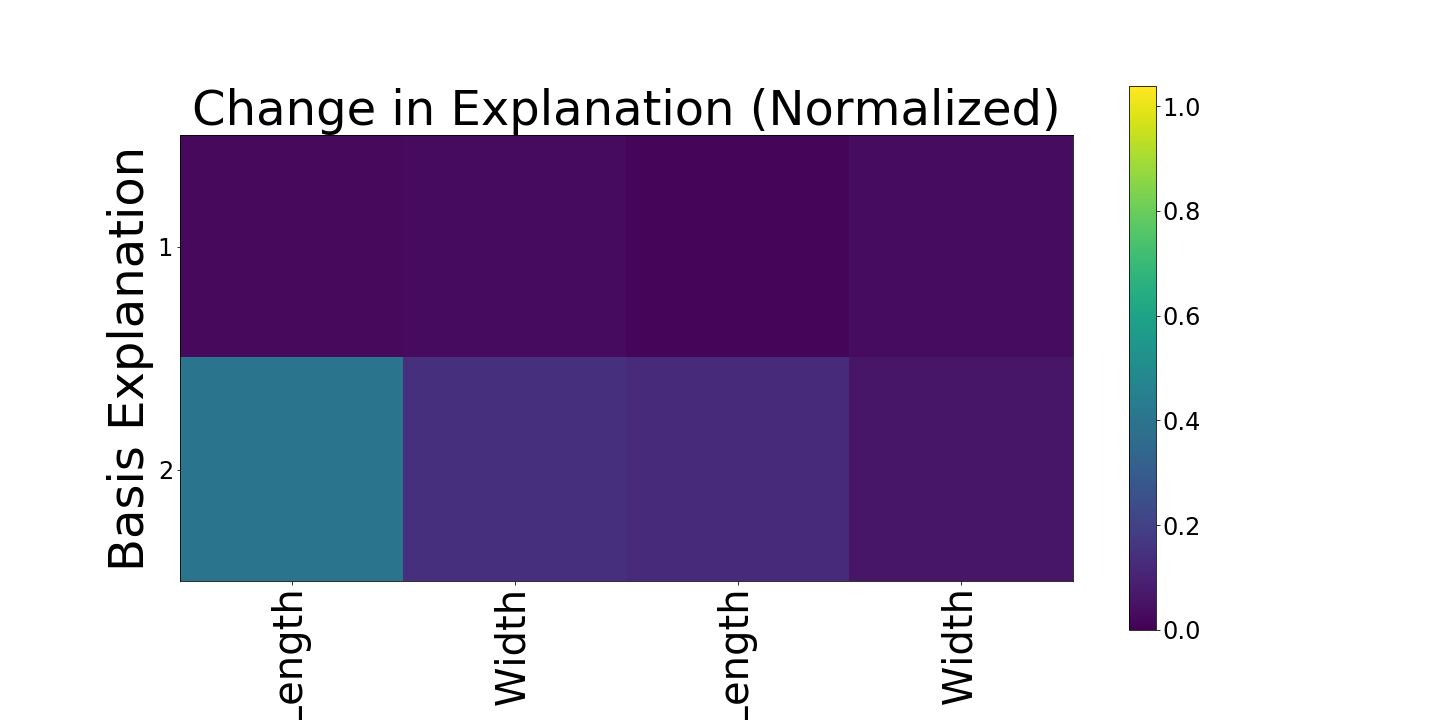} 
\end{minipage}%
\hspace{8pt}
\begin{minipage}{.45\linewidth}
	\centering
	\includegraphics[width = \linewidth]{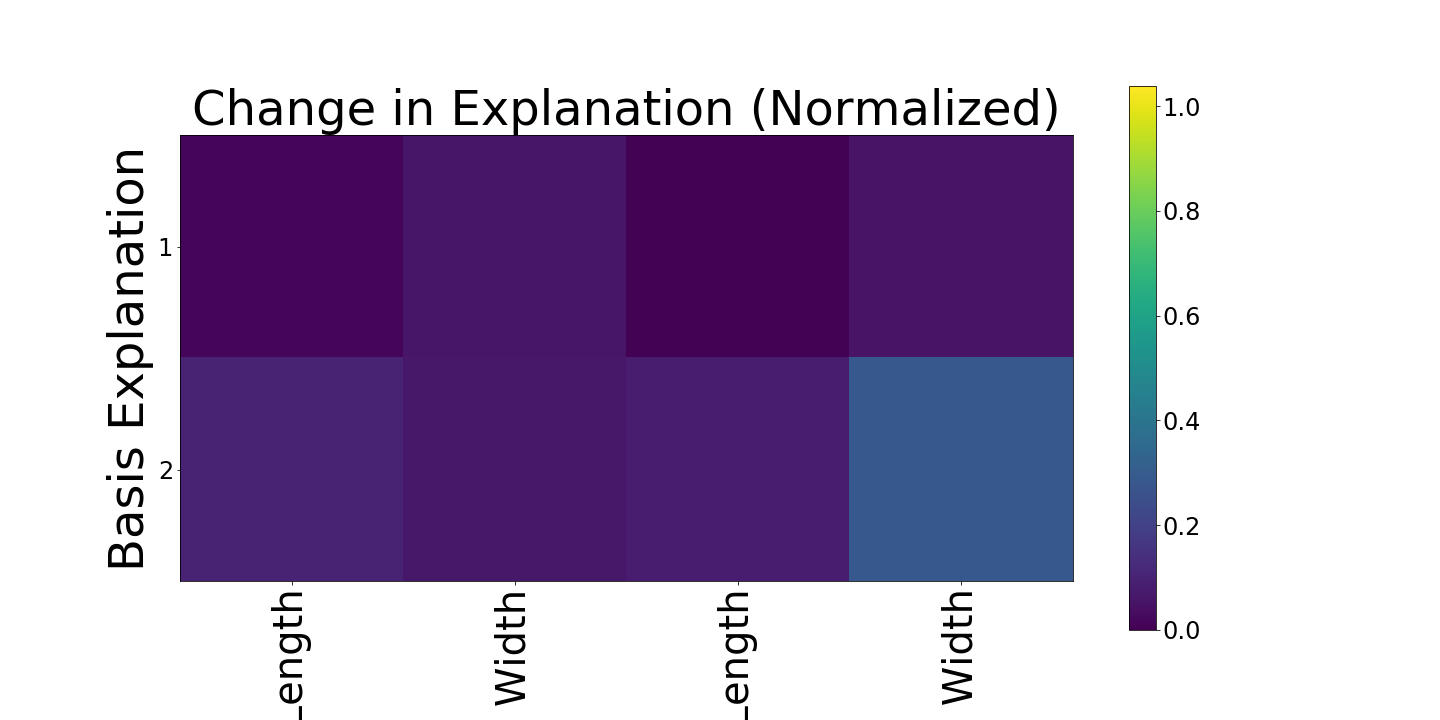}
\end{minipage}
\captionof{figure}{The absolute difference between the Modified (Left)/Retrained (Right) explanations and the Original explanations scaled relative to the Original explanations on the Iris dataset.}
\label{fig:a-iris-similarity}

\begin{minipage}{.45\linewidth}
	\centering
	\includegraphics[width = \linewidth]{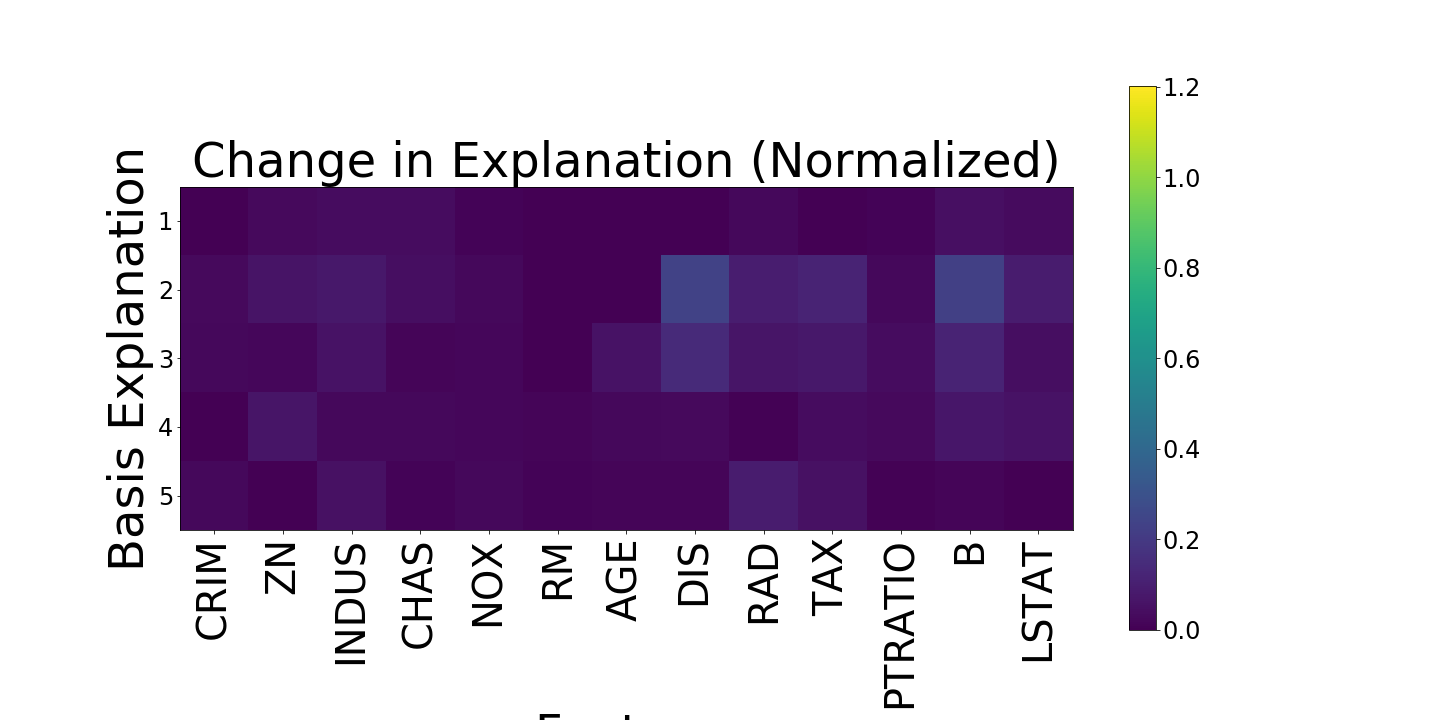} 
\end{minipage}%
\hspace{8pt}
\begin{minipage}{.45\linewidth}
	\centering
	\includegraphics[width = \linewidth]{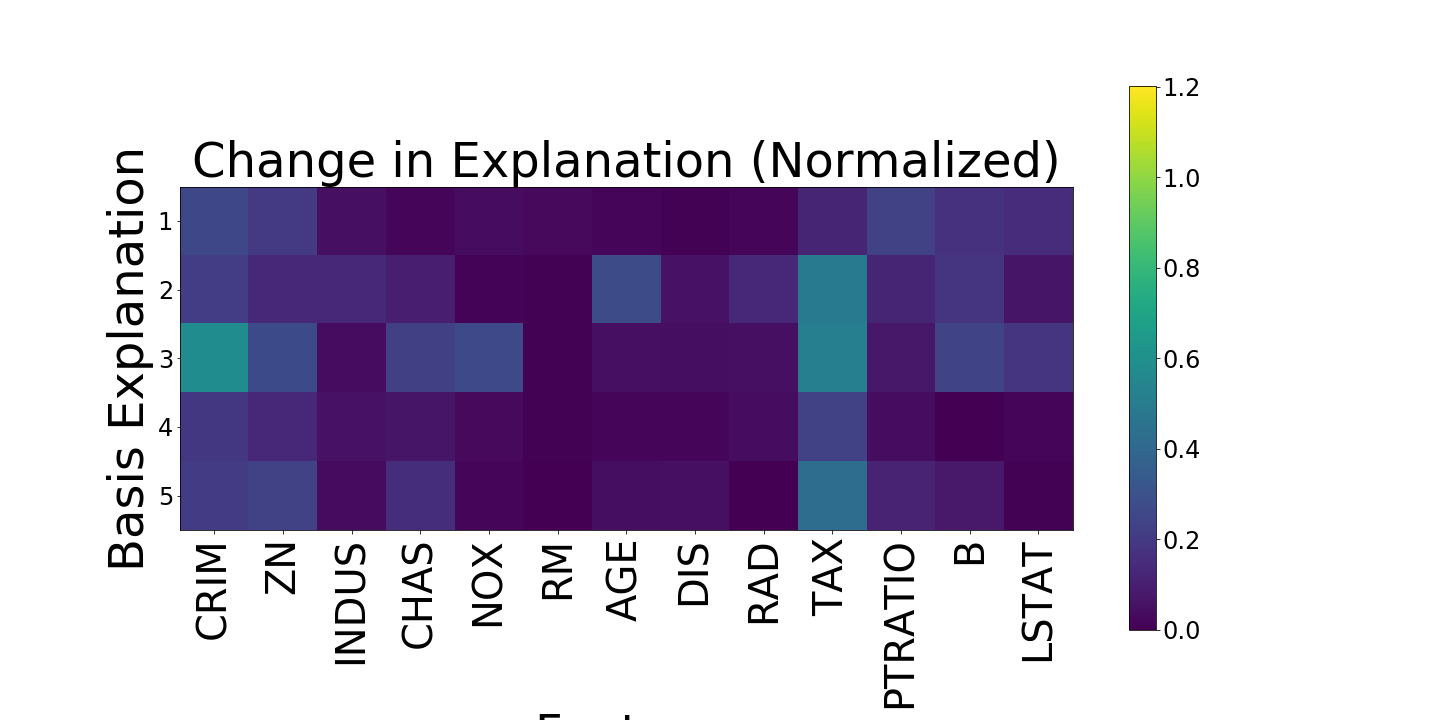}
\end{minipage}
\captionof{figure}{The absolute difference between the Modified (Left)/Retrained (Right) explanations and the Original explanations scaled relative to the Original explanations on the Boston Housing dataset.}
\label{fig:a-housing-similarity}

\begin{minipage}{.45\linewidth}
	\centering
	\includegraphics[width = \linewidth]{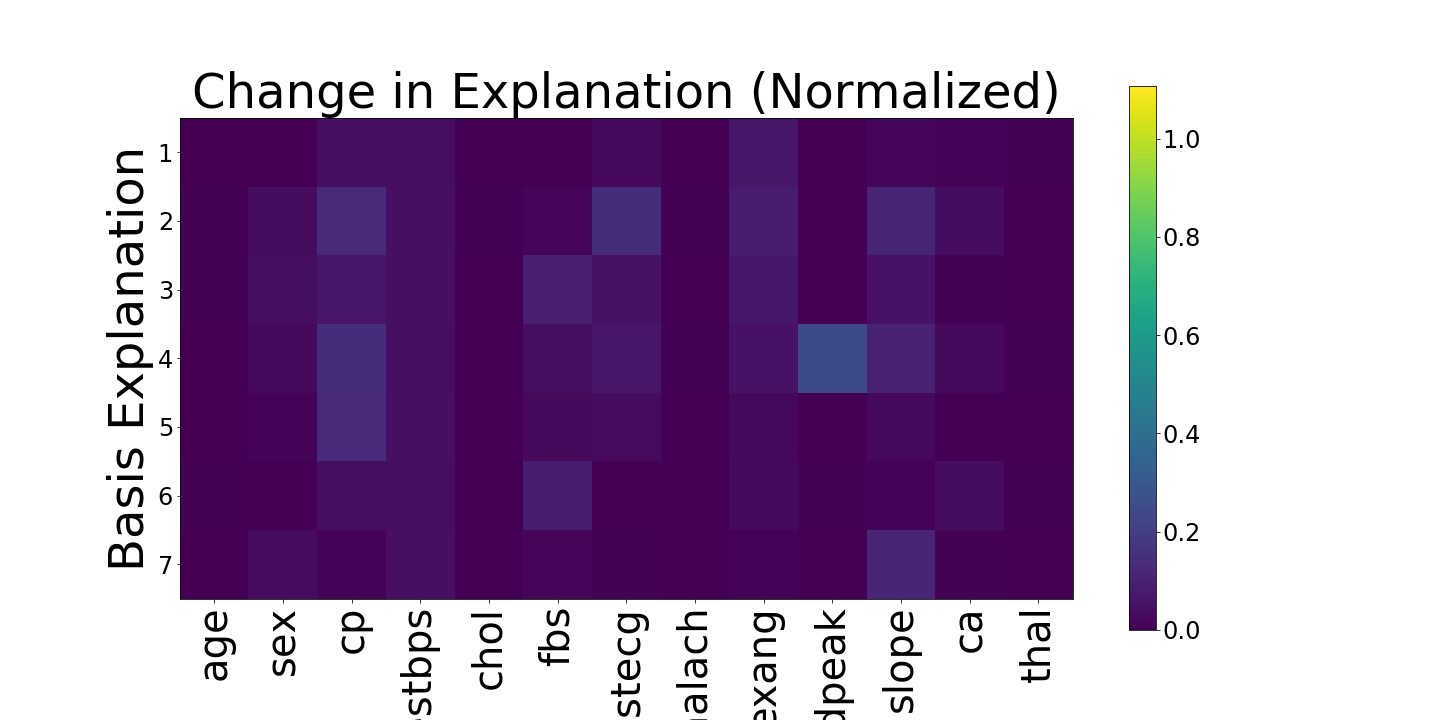} 
\end{minipage}%
\hspace{8pt}
\begin{minipage}{.45\linewidth}
	\centering
	\includegraphics[width = \linewidth]{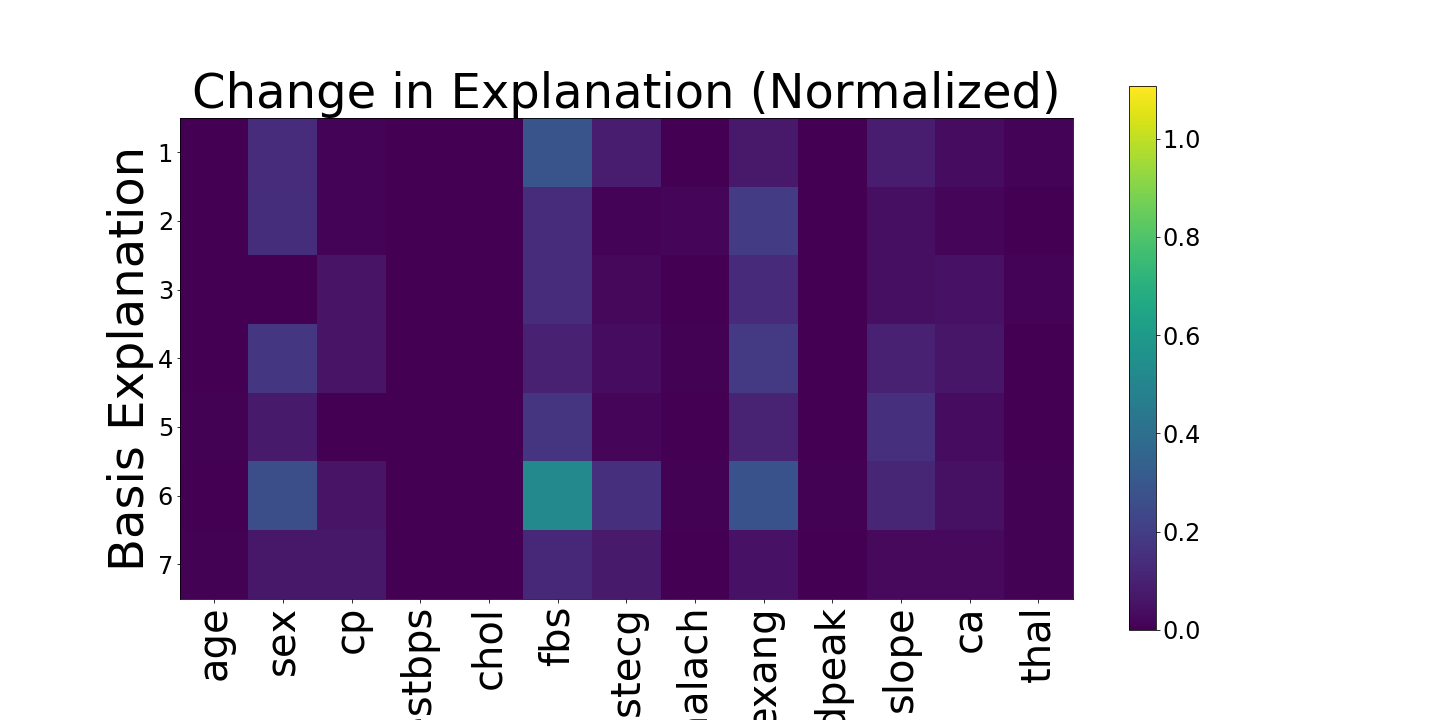}
\end{minipage}
\captionof{figure}{The absolute difference between the Modified (Left)/Retrained (Right) explanations and the Original explanations scaled relative to the Original explanations on the Heart Disease dataset.}
\label{fig:a-heart-similarity}

\end{document}